\documentclass{article}

\PassOptionsToPackage{round}{natbib}
\usepackage[preprint]{neurips_2026}

\usepackage[utf8]{inputenc} \usepackage[T1]{fontenc}    \usepackage{hyperref}       \usepackage{url}            \usepackage{booktabs}       \usepackage{amsfonts}       \usepackage{nicefrac}       \usepackage{microtype}      \usepackage{xcolor}         \usepackage{graphicx} \usepackage{amsmath}
\usepackage{subcaption}
\usepackage{listings}
\usepackage{upquote}
\usepackage{float}
\usepackage{tabularx}
\usepackage{titletoc}
\newcolumntype{Y}{>{\raggedright\arraybackslash}X}

\definecolor{codegreen}{rgb}{0,0.6,0}
\definecolor{codegray}{rgb}{0.5,0.5,0.5}
\definecolor{codepurple}{rgb}{0.58,0,0.82}
\definecolor{backcolour}{rgb}{0.95,0.95,0.92}

\lstdefinestyle{pythonstyle}{
    backgroundcolor=\color{backcolour},   
    commentstyle=\color{codegreen},
    keywordstyle=\color{magenta},
    numberstyle=\tiny\color{codegray},
    stringstyle=\color{codepurple},
    basicstyle=\ttfamily\footnotesize,
    breakatwhitespace=false,         
    breaklines=true,                 
    captionpos=b,                    
    keepspaces=true,                 
    numbers=none,                    
    numbersep=5pt,                  
    showspaces=false,                
    showstringspaces=false,
    showtabs=false,                  
    tabsize=2
}

\lstdefinestyle{cppstyle}{
    backgroundcolor=\color{backcolour},   
    commentstyle=\color{codegreen},
    keywordstyle=\color{blue},
    numberstyle=\tiny\color{codegray},
    stringstyle=\color{codepurple},
    basicstyle=\ttfamily\footnotesize,
    breakatwhitespace=false,         
    breaklines=true,                 
    captionpos=b,                    
    keepspaces=true,                 
    numbers=none,                    
    numbersep=5pt,                  
    showspaces=false,                
    showstringspaces=false,
    showtabs=false,                  
    tabsize=2
}

\lstdefinestyle{csharpstyle}{
    backgroundcolor=\color{backcolour},   
    commentstyle=\color{codegreen},
    keywordstyle=\color{blue},
    numberstyle=\tiny\color{codegray},
    stringstyle=\color{codepurple},
    basicstyle=\ttfamily\footnotesize,
    breakatwhitespace=false,         
    breaklines=true,                 
    captionpos=b,                    
    keepspaces=true,                 
    numbers=none,                    
    numbersep=5pt,                  
    showspaces=false,                
    showstringspaces=false,
    showtabs=false,                  
    tabsize=2,
    language=[Sharp]C
}

\lstdefinestyle{javastyle}{
    backgroundcolor=\color{backcolour},   
    commentstyle=\color{codegreen},
    keywordstyle=\color{blue},
    numberstyle=\tiny\color{codegray},
    stringstyle=\color{codepurple},
    basicstyle=\ttfamily\footnotesize,
    breakatwhitespace=false,         
    breaklines=true,                 
    captionpos=b,                    
    keepspaces=true,                 
    numbers=none,                    
    numbersep=5pt,                  
    showspaces=false,                
    showstringspaces=false,
    showtabs=false,                  
    tabsize=2,
    language=Java
}

\lstdefinestyle{jsonstyle}{
    backgroundcolor=\color{backcolour},
    commentstyle=\color{codegreen},
    keywordstyle=\color{blue},
    stringstyle=\color{codepurple},
    basicstyle=\ttfamily\footnotesize,
    breakatwhitespace=false,
    breaklines=true,
    captionpos=b,
    keepspaces=true,
    numbers=none,
    numbersep=5pt,
    showspaces=false,
    showstringspaces=false,
    showtabs=false,
    tabsize=2,
    literate=
        *{:}{{{\color{blue}{:}}}}{1}
        {,}{{{\color{blue}{,}}}}{1}
        {\{}{{{\color{blue}{\{}}}}{1}
        {\}}{{{\color{blue}{\}}}}}{1}
        {[}{{{\color{blue}{[}}}}{1}
        {]}{{{\color{blue}{]}}}}{1},
    morestring=[b]",
    morecomment=[s]{/*}{*/},
    sensitive=true
}

\lstdefinelanguage{JavaScript}{
  keywords={typeof, new, true, false, catch, function, return, null, catch, switch, var, if, in, while, do, else, case, break},
  keywordstyle=\color{blue}\bfseries,
  ndkeywords={class, export, boolean, throw, implements, import, this},
  ndkeywordstyle=\color{darkgray}\bfseries,
  identifierstyle=\color{black},
  sensitive=false,
  comment=[l]{//},
  morecomment=[s]{/*}{*/},
  commentstyle=\color{purple}\ttfamily,
  stringstyle=\color{red}\ttfamily,
  morestring=[b]',
  morestring=[b]"
}

\lstdefinestyle{javascriptstyle}{
    backgroundcolor=\color{backcolour},
    commentstyle=\color{codegreen},
    keywordstyle=\color{blue},
    ndkeywordstyle=\color{darkgray}\bfseries,
    numberstyle=\tiny\color{codegray},
    stringstyle=\color{codepurple},
    basicstyle=\ttfamily\footnotesize,
    breakatwhitespace=false,
    breaklines=true,
    captionpos=b,
    keepspaces=true,
    columns=fullflexible,
    numbers=none,
    numbersep=5pt,
    showspaces=false,
    showstringspaces=false,
    showtabs=false,
    tabsize=2,
    language=JavaScript
}

\lstdefinelanguage{TypeScript}{
  keywords={abstract, break, case, catch, class, const, continue, debugger, default, delete, do, else, enum, export, extends, false, finally, for, function, if, import, in, instanceof, new, null, return, super, switch, this, throw, true, try, typeof, var, void, while, with, let, static, readonly, private, protected, public, implements, interface, package, yield, async, await},
  keywordstyle=\color{blue}\bfseries,
  ndkeywords={any, boolean, number, string, undefined, never, unknown, Symbol, Array, ReadonlyArray, object, type, namespace, module},
  ndkeywordstyle=\color{darkgray}\bfseries,
  identifierstyle=\color{black},
  sensitive=false,
  comment=[l]{//},
  morecomment=[s]{/*}{*/},
  commentstyle=\color{codegreen}\ttfamily,
  stringstyle=\color{codepurple}\ttfamily,
  morestring=[b]',
  morestring=[b]",
  morestring=[b]`
}

\lstdefinestyle{typescriptstyle}{
    backgroundcolor=\color{backcolour},
    commentstyle=\color{codegreen},
    keywordstyle=\color{blue},
    ndkeywordstyle=\color{darkgray}\bfseries,
    numberstyle=\tiny\color{codegray},
    stringstyle=\color{codepurple},
    basicstyle=\ttfamily\footnotesize,
    breakatwhitespace=false,
    breaklines=true,
    captionpos=b,
    keepspaces=true,
    columns=fullflexible,
    numbers=none,
    numbersep=5pt,
    showspaces=false,
    showstringspaces=false,
    showtabs=false,
    tabsize=2,
    language=TypeScript
}

\definecolor{goldenbackground}{rgb}{0.9,1.0,0.9}
\definecolor{modelbackground}{rgb}{0.95,0.95,1.0}
\definecolor{prefixbackground}{rgb}{0.98,0.98,0.98}
\definecolor{suffixbackground}{rgb}{0.98,0.98,0.98}
\definecolor{assertionbackground}{rgb}{0.98,0.98,0.98}

\lstdefinestyle{golden}{
    backgroundcolor=\color{goldenbackground},
    basicstyle=\ttfamily\footnotesize,
    breaklines=true,
    frame=single,
    framesep=2pt,
    rulecolor=\color{green!40!black},
    title=\textbf{Golden Completion}
}

\newcommand{\modelname}{Claude}
\lstdefinestyle{model}{
    backgroundcolor=\color{modelbackground},
    basicstyle=\ttfamily\footnotesize,
    breaklines=true,
    frame=single,
    framesep=2pt,
    rulecolor=\color{blue!40!black},
    title=\textbf{Model Completion (\modelname)}
}

\lstdefinestyle{prefix}{
    backgroundcolor=\color{prefixbackground},
    basicstyle=\ttfamily\footnotesize,
    breaklines=true,
    frame=tb,
    framesep=2pt,
    title=\textbf{Prefix}
}

\lstdefinestyle{suffix}{
    backgroundcolor=\color{suffixbackground},
    basicstyle=\ttfamily\footnotesize,
    breaklines=true,
    frame=tb,
    framesep=2pt,
    title=\textbf{Suffix}
}

\lstdefinestyle{assertion}{
    backgroundcolor=\color{assertionbackground},
    basicstyle=\ttfamily\footnotesize,
    breaklines=true,
    frame=tb,
    framesep=2pt,
    title=\textbf{Assertions}
}

\title{DevBench: A Realistic, Developer-Informed Benchmark for Code Generation Models}

\author{
  \textbf{Adarsh Kumarappan\thanks{Work done during an internship at Microsoft.}\hspace{0.4em}$^{,1}$,
  Pareesa Ameneh Golnari$^{2}$,
  Wen Wen$^{2}$,
  Xiaoyu Liu$^{2}$,} \\
  \textbf{Gabriel Ryan$^{2}$,
  Yuting Sun$^{2}$,
  Shengyu Fu$^{2}$,
  Elsie Nallipogu$^{2}$} \\[0.3em]
  $^{1}$California Institute of Technology,
  $^{2}$Microsoft \\
  \texttt{adarsh@caltech.edu, \{pgolnar, wenwen, lixiaoyu, ryangabriel,} \\
  \texttt{yutingsun, shengyfu, elsien\}@microsoft.com}
}

\begin{document}

\maketitle

\begin{abstract}
\textbf{DevBench} is a telemetry-driven benchmark designed to evaluate Large Language Models (LLMs) on realistic code completion tasks. It includes 1,800 evaluation instances across six programming languages and six task categories derived from real developer telemetry and synthesized using generator models from multiple provider families to mitigate single-source bias. Unlike prior benchmarks, it emphasizes ecological validity, avoids training data contamination, and enables detailed diagnostics. The evaluation combines functional correctness, similarity-based metrics, and LLM-judge assessments focused on usefulness and contextual relevance. 9 state-of-the-art models were assessed, with the strongest achieving only 43.5\% Pass@1, confirming the benchmark remains challenging and revealing differences in syntactic precision, semantic reasoning, and practical utility. Our benchmark provides actionable insights to guide model selection and improvement, detail that is often missing from other benchmarks but is essential for both practical deployment and targeted model development.
\end{abstract}
 \section{Introduction}
\label{sec:intro}

Large Language Models (LLMs) have transformed modern software development by enabling advanced code generation, powering tools like GitHub Copilot~\citep{github_copilot_2025} and Cursor~\citep{cursor_2025}. As these systems are increasingly integrated into real-world workflows, realistic and rigorous evaluation frameworks are essential to understanding their strengths and limitations.

Existing benchmarks evaluate different aspects of code generation: problem solving benchmarks for coding problems~\citep{chenEvaluatingLargeLanguage2021,austinProgramSynthesisLarge2021,hendrycksMeasuringCodingChallenge2021,iyerMappingLanguageCode2018,yinLearningMineAligned2018}, repository-based benchmarks for challenges in large projects~\citep{wuRepoMasterEvalEvaluatingCode2024,duClassEvalManuallyCraftedBenchmark2023, dingCrossCodeEvalDiverseMultilingual2023,yuCoderEvalBenchmarkPragmatic2024,zhuoBigCodeBenchBenchmarkingCode2025,jimenezSWEbenchCanLanguage2024,dengSWEBenchProCan2025}, and evolving benchmarks addressing contamination~\citep{liEvoCodeBenchEvolvingCode2024, jainLiveCodeBenchHolisticContamination2024}.

However, existing benchmarks rely on code samples scraped from open source repositories or coding challenge websites and generate target completions based on static rules for filling in line, function, or class implementations. This limits them in several ways: First, the target completions are not based on real world  usage patterns for code completion tools, and therefore do not focus on common challenging completion scenarios that arise in real world usage. Second, the diagnostic value of these benchmarks is limited because they report aggregate metrics, but cannot attribute differences in performance to specific usage areas. Third, benchmarks collected from publicly available sources are prone to training data contamination, which has been observed in models overfitting to existing benchmarks~\citep{jainLiveCodeBenchHolisticContamination2024}.

To address these limitations, we introduce \textbf{DevBench}, a realistic and scalable benchmark grounded in observed developer behavior. \textbf{DevBench} focuses on common yet challenging completion scenarios, identified from internal telemetry of over \textbf{one billion developer code completion interactions} and synthesized into 1,800 evaluation instances spanning six languages and six empirically distinct task categories (mean pairwise $\bar{\rho}{=}0.44$), using generator models from multiple provider families with validated cross-family fairness ($\rho{=}0.91$ ranking consistency). Each instance is reviewed for quality and realism, ensuring that tasks reflect how developers actually use code completion tools while remaining contamination-resistant.

As shown in Table~\ref{tab:benchmark-comparison}, \textbf{DevBench} advances beyond existing benchmarks in both realism~\citep{paulBenchmarksMetricsEvaluations2024} and scope. It offers four key advantages: \textbf{(1) realism}, with tasks rooted in observed developer behavior; \textbf{(2) contamination resistance}, through synthetic but controlled instance generation; \textbf{(3) fine-grained evaluation}, assessing semantic alignment and developer utility; and \textbf{(4) cross-language coverage}, spanning Python (Py), JavaScript (JS), TypeScript (TS), Java, C++, and C\#.

Together, these features provide ecological validity: \textbf{DevBench} reflects \textbf{authentic developer challenges} rather than hypothetical tasks, is validated through expert review, and captures diverse contexts across languages and developer skill levels. By enabling both overall rankings and scenario-specific diagnostics, \textbf{DevBench} supports informed model selection and optimization, and provides a contamination-resilient foundation for future research. We open-source the 1,800-instance benchmark and evaluation code.\footnote{Code and benchmark: \url{https://github.com/microsoft/devbench}}

\begin{table*}[t]
  \caption{Comparison of recent code generation benchmarks across size, language coverage, focus, source, and unique features.}
  \label{tab:benchmark-comparison}
  \centering
  \scriptsize
  \setlength{\tabcolsep}{3pt}
  \renewcommand{\arraystretch}{0.95}
  \begin{tabular}{@{}p{1.9cm}p{0.8cm}p{1.8cm}p{2.2cm}p{2.2cm}p{2.7cm}@{}}
    \toprule
    \textbf{Benchmark} & \textbf{\# Tasks} & \textbf{Languages} & \textbf{Focus} & \textbf{Source} & \textbf{Unique Feature} \\
    \midrule
    RepoMasterEval & 288  & Py, TS & Real-world repo completion & GitHub repos ({>}100 stars) & Mutation testing for test robustness \\
    CrossCodeEval & ${\sim}$10k & Py, Java, TS, C\# & Cross-file dependencies & GitHub repos ({>}3 stars) & Static analysis for dependencies \\
    CoderEval & 460  & Py, Java & Cross-file pragmatic generation & GitHub repos (popular tags) & Human-labeled docstrings \\
    ClassEval & 100  & Py  & Class-level generation & Manually crafted & Multiple interdependent methods \\
    HumanEval & 164  & Py  & Basic programming tasks & Manually crafted & Simple interview-style problems \\
    HumanEval+ & 164  & Py  & Enhanced testing rigor & Manually crafted & 80$\times$ more evaluation instances \\
    LiveCodeBench & 511  & Py & Contamination-free evaluation & Competition platforms & Time-based contamination tracking \\
    SWE-bench & 2,294  & Py & Repo-level bug fixing & GitHub issues and PRs & Real-world issues from 12 popular repos \\
    BigCodeBench & 1,140  & Py & Diverse function calls as tools & Human-LLM collab.\ generation & 723 function calls from 139 libs \\
    \midrule
    \textbf{DevBench (ours)} & \textbf{1,800} & \textbf{Py, JS, TS, Java, C++, C\#} & \textbf{Realistic developer-informed scenarios} & \textbf{Synth.\ generated, manually reviewed} & \textbf{Telemetry-guided, human-validated} \\
    \bottomrule
  \end{tabular}
\end{table*}

 \section{Benchmark Design}
\label{sec:benchmark}

We view code generation as a composite, puzzle-solving task in which models must combine distinct capabilities, such as API usage, intent understanding, and syntax control. To evaluate these skills, we define benchmark categories that isolate each capability while ensuring every instance is solvable from the provided prefix/suffix, making evaluation both realistic and fair. Although individual instances are synthesized, \textbf{DevBench} is telemetry-driven: its categories, task types, and scenarios are derived from analysis of over one billion real developer interactions, with synthesis used only to instantiate these empirically derived patterns in a privacy-preserving, contamination-resistant manner.

\begin{figure}[t]
    \centering
    \vspace{-1em}
    \includegraphics[width=0.99\linewidth,trim=0 25 0 25,clip]{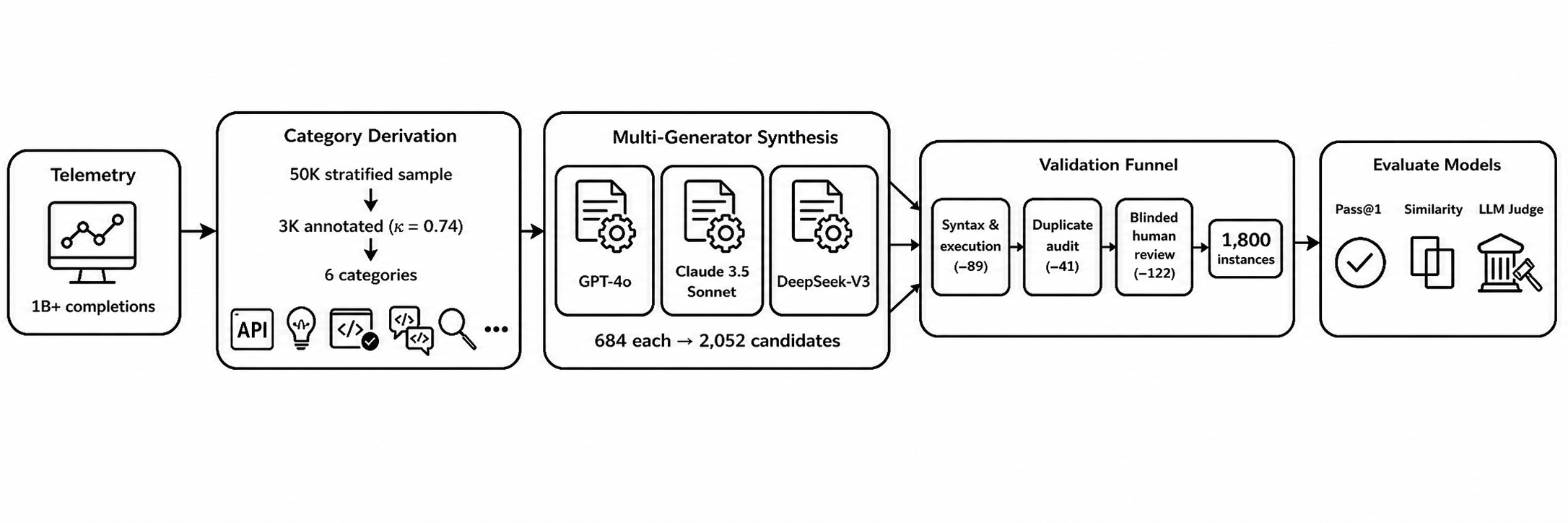}
    \vspace{-1.5em}
    \caption{End-to-end \textbf{DevBench} pipeline. Telemetry analysis of 1B+ completions produces six task categories via stratified sampling, annotation ($\kappa = 0.74$), and clustering. Three generator models from different provider families (GPT-4o, Claude 3.5 Sonnet, DeepSeek-V3) each produce 684 candidate instances. A three-stage validation funnel (automatic syntax/execution checks, near-duplicate filtering against public benchmarks, and blinded human review) reduces 2,052 candidates to the final 1,800 instances. Models are evaluated on functional correctness (Pass@1), similarity metrics, and LLM-judge assessments.}
    \label{fig:pipeline}
    \vspace{-0.5em}
\end{figure}

\subsection{From User Telemetry to Categories}
\label{subsec:telemetry_to_categories}

\textbf{DevBench} is derived from an internal telemetry corpus of over one billion anonymized code-completion interactions, including prefix/suffix context, generated completions and final accepted completions, and user actions (accept, reject, edit). This dataset spans diverse contexts over IDEs, geographical locations, language distribution, and developers ranging from students to senior engineers. To satisfy privacy and compliance requirements, we do not release or reuse raw user code; all benchmark instances are synthetic.

Category derivation used a three-stage telemetry analysis (Appendix~\ref{sec:telemetry_derivation}). We first drew a stratified, difficulty-enriched sample of 50,000 completions, balanced across the six target languages and enriched for low-acceptance or high-edit-distance cases. Three researchers then annotated 3,000 completions (500 per language) with scenario and failure-mode labels, achieving substantial agreement on the primary label (Fleiss' $\kappa = 0.74$ before adjudication). The resulting clusters were consolidated into six task categories retained only if they satisfied \emph{prevalence} (appearing in $\geq$5\% of low-acceptance completions across at least four languages) and \emph{actionability} (isolable into a self-contained prefix/suffix completion task). Because the telemetry distribution is not uniform across languages, the released benchmark is balanced rather than proportional: 300 instances per language and 50 per language--category cell, prioritizing statistical power for cross-language diagnostics. \textbf{DevBench} is therefore a telemetry-grounded, balanced diagnostic benchmark rather than a proportional sample of all completion traffic.

\subsection{Benchmark categories}
\label{subsec:benchmark_categories}

We define six benchmark categories based on our analysis of user telemetry. Each category targets a distinct type of developer intent and is consistently evaluated across languages, with adaptations to reflect the idioms and ecosystems of each target language (see Table~\ref{tab:language-adaptations}). The categories are described in detail below (examples in Appendix~\ref{sec:category_examples}).

\textbf {1.\ API Usage}:
This category tests a model's ability to correctly apply specialized library functions.
Each evaluation instance consists of a prefix that sets the context, a golden completion illustrating proper API usage, and a suffix for continuation.

\textbf{2.\ Code Purpose Understanding}: This category evaluates whether a model can infer business logic, domain conventions, and state constraints from surrounding code, rather than merely producing syntactically valid completions. For instance, given a \texttt{Subscription} class with state transitions, an audit log, and a configurable cooldown period, the model must implement a \texttt{resume} method that checks cooldown elapsed time, verifies the current state is \texttt{paused}, and reuses the existing \texttt{\_transition} helper for audit-log consistency.

\begin{table*}[t]
  \caption{Language-specific adaptations. NL = Natural Language, HOFs = Higher-Order Functions, RAII = Resource Acquisition Is Initialization.}
  \label{tab:language-adaptations}
  \centering
  \scriptsize
  \renewcommand{\arraystretch}{0.95}
  \setlength{\tabcolsep}{3pt}
  \begin{tabularx}{\textwidth}{@{}>{\raggedright\arraybackslash}p{1.5cm} Y Y Y Y Y Y@{}}
    \toprule
    \textbf{Category} & \textbf{Python} & \textbf{C\#} & \textbf{C++} & \textbf{Java} & \textbf{JavaScript} & \textbf{TypeScript} \\
    \midrule
    API Usage & inspect, weakref, csv dialects & LINQ, Span, Reflection & Smart ptrs, concurrency, charconv & DateTimeFormatter, ConcurrentHashMap & Node.js crypto, Buffer, Atomics & Same as JS w/ types \\
    \cmidrule{1-7}
    Code Purpose & Iterators/ generators, context mgrs & LINQ/ collections, decimal arith. & Containers, struct workflows & Streams/ collections, BigDecimal & Closures, class state machines & Same as JS w/ type systems \\
    \cmidrule{1-7}
    Code2NL & Docstrings & XML docs & Doxygen comments & Javadoc & JSDoc & TSDoc w/ type annot. \\
    \cmidrule{1-7}
    Low Context & Decorators, context mgrs & Span, LINQ, yield & RAII, move semantics & Collections views, atomics & Proxy, WeakMap, HOFs & Type guards, branded types \\
    \cmidrule{1-7}
    Pattern Match. & Config parsers, template dispatch & Config parsers, edit distance & Line-continuation, bracket matching & Line-continuation, check digits & Template engines, deep equality & Escape-aware parsers, tokenizers \\
    \cmidrule{1-7}
    Syntax Compl. & Match/case, metaclasses, async & Switch exprs, records, LINQ & Fold exprs, constexpr if, RAII & Sealed classes, records, switch patterns & Async generators, Proxy, Symbols & Conditional types, overloads \\
    \bottomrule
  \end{tabularx}
\end{table*}

\textbf{3.\ Code2NL/NL2Code}:
This category evaluates a model's ability to translate between code and natural language (NL) in both directions. This reflects real-world developer workflows, where boundaries between code and language are increasingly blurred. To align with practical use cases, our benchmark covers a wide spectrum of scenarios including: (1) NL only in the prefix, (2) code only in the prefix, (3) mixed NL and code in the prefix, (4) various NL forms including docstrings, inline comments, block comments, and user-facing documentation, and (5) different documentation styles across programming languages (e.g., Python docstrings, JavaDoc, JSDoc, XML docs, and Doxygen).

\textbf{4.\ Low Context}:
This category evaluates a model's ability to complete code using minimal context (10--20 lines total), requiring it to recognize language-specific patterns and idioms. These tasks are carefully designed to be solvable despite limited information, testing the model's deep understanding of programming conventions without relying on broader context.

\textbf{5.\ Pattern Matching}:
This category tests a model's ability to recognize and extend established code patterns within realistic contexts. The prefix establishes a pattern through examples or specifications, and the model must implement a structurally similar but semantically distinct continuation, following the intended structure rather than generating arbitrary code.

\textbf{6.\ Syntax Completion}:
This category tests a model's ability to generate complex, nested structures while adhering to language-specific syntax rules. Evaluation instances span four categories: nested control structures, complex features, multi-line patterns, and error handling. The model must correctly manage indentation, close code blocks, and match braces or parentheses, demonstrating mastery of each language's unique syntactic constructs.

\textbf{Category relationships.} Because real developer tasks blend capabilities, we do not expect perfect category orthogonality. Empirically, pairwise Spearman correlations of Pass@1 across 54 model--language combinations have mean $\bar{\rho}=0.44$, with low-correlation pairs such as API Usage vs.\ Code Purpose Understanding ($\rho=0.16$) and API Usage vs.\ Pattern Matching ($\rho=0.17$). Conversely, Pattern Matching and Code Purpose Understanding correlate more highly ($\rho=0.74$), yet still diverge diagnostically for individual models (e.g., DeepSeek V4 Pro excels at Pattern Matching but underperforms GPT-5.5 on Syntax Completion, 32.4\% vs.\ 51.4\%; Table~\ref{tab:model-performance}). PCA shows the first two components explain $\approx$77\% of variance and three explain $\approx$87\%, indicating two to three capability dimensions beyond a single code-generation factor. Full correlations are in Appendix~\ref{sec:category_correlations}.

\subsection{Benchmark construction}
\label{subsec:benchmark_construction}
Figure~\ref{fig:pipeline} overviews the construction pipeline: starting from telemetry-derived categories, we generate the benchmark.

\textbf{Evaluation Instance Structure}: Each instance contains a prefix, a golden completion, an optional suffix for fill-in-the-middle (FIM) settings, and hidden assertions for functional validation. Assertions are never visible to the evaluated model, and cursor positions are chosen at natural code boundaries observed in telemetry, covering both prefix-only and FIM-style completions.

\textbf{Generation and Validation}: To mitigate single-generator stylistic bias, instances were synthesized using three generator models from different provider families: GPT-4o~\citep{openaiGPT4oSystemCard2024}, Claude~3.5 Sonnet~\citep{IntroducingClaude35}, and DeepSeek-V3~\citep{deepseek-aiDeepSeekV3TechnicalReport2025}. These generators are earlier than the corresponding evaluated frontier models, reducing circularity between synthesis and evaluation. Each generator produced 684 candidates using the same structured prompts (Appendix~\ref{subsec:benchmark_generation_prompts}), yielding 2,052 candidates. After automatic syntax/execution validation, near-duplicate filtering against public benchmarks, and blinded human review, the final benchmark contains 1,800 instances with near-balanced generator provenance (GPT-4o: 597, Claude~3.5 Sonnet: 601, DeepSeek-V3: 602). Each instance was independently reviewed by two annotators from a team of three senior researchers and engineers with expertise across all six target languages, evaluating four dimensions: (1) \emph{usefulness} (plausible developer need), (2) \emph{realism} (authentic coding patterns, including common inconsistencies), (3) \emph{category alignment} (consistent with intended task type), and (4) \emph{complexity authenticity} (genuine difficulty observed in telemetry). Inter-annotator agreement was substantial (Cohen's $\kappa=0.82$, 91\% raw agreement); the most common rejection reasons were overly simplified or ``textbook-perfect'' implementations (32\%), insufficient complexity (28\%), unrealistic edge-case handling (23\%), and category misalignment (17\%). Full validation details are in Appendix~\ref{sec:generation_validation_details}.

\textbf{Complexity and Diversity}: \textbf{DevBench} instances average 57.6 LOC with a balanced prefix/completion structure (117.9 prefix tokens, 108.2 completion tokens), as shown in Tables~\ref{tab:benchmark-complexity} and~\ref{tab:language-stats}. In contrast, CrossCodeEval features long prompts (71--116 LOC) but extremely short completions (1--2 LOC); \textbf{DevBench}'s balance is more reflective of practical code-completion workflows. To verify that repeated use of the same language--category prompt did not produce homogeneous instances, we measured within-prompt Term Frequency--Inverse Document Frequency (TF-IDF) cosine similarity; the mean pairwise similarity is 0.094 (cosine distance 0.906), indicating high within-prompt diversity (Appendix~\ref{sec:within_prompt_diversity}).

\textbf{Generator Bias Analysis.}
\label{subsec:generator_bias_analysis}
We evaluated generator-family favoritism by fitting a logistic regression over model--instance pass/fail outcomes with fixed effects for evaluated model, generator family, language, category, and completion-length bin. The same-family coefficient was small and not statistically significant, and model rankings on each generator's subset were highly correlated (minimum pairwise Spearman $\rho=0.91$). We find no evidence of systematic generator-family favoritism. To reduce instance-level contamination risk, we also filtered candidates against the established benchmarks in Table~\ref{tab:benchmark-complexity} using token 4-gram Jaccard and AST structural similarity, removing 41 of 2,052 candidates (2.0\%). \textbf{DevBench} is thus designed to be contamination-resistant at the instance level while reflecting general programming patterns that all code models are expected to learn; full methodology is in Appendix~\ref{sec:generator_bias_details}.

\begin{table*}[t]
\scriptsize
\centering
\begin{minipage}[t]{0.34\textwidth}
\centering
\caption{Avg.\ LOC of code generation benchmarks.}
\label{tab:benchmark-complexity}
\renewcommand{\arraystretch}{0.95}
\begin{tabular*}{\linewidth}{@{\extracolsep{\fill}}lr@{}}
\toprule
\textbf{Benchmark} & \textbf{Avg.\ LOC} \\
\midrule
\textbf{DevBench (ours)} & 57.6 \\
CrossCodeEval & 71--117 \\
CoderEval-Py & 32.0 \\
APPS & 21.4 \\
HumanEval & 11.5 \\
CoderEval-Java & 10.2 \\
MBPP & 6.8 \\
Concode & 4.8 \\
DS-1000 & 3.8 \\
CoNaLA & 1.0 \\
\bottomrule
\end{tabular*}
\end{minipage}%
\hfill
\begin{minipage}[t]{0.62\textwidth}
\centering
\caption{\textbf{DevBench} language-specific statistics.}
\label{tab:language-stats}
\renewcommand{\arraystretch}{0.95}
\resizebox{\linewidth}{!}{%
\begin{tabular}{@{}lrrrrr@{}}
\toprule
\textbf{Language} & \textbf{Prefix LOC} & \textbf{Compl.\ LOC} & \textbf{Total LOC} & \textbf{Prefix Tok.} & \textbf{Compl.\ Tok.} \\
\midrule
Python & 14.7 & 11.3 & 48.8 & 71.6 & 81.3 \\
C\# & 20.2 & 12.4 & 58.5 & 121.3 & 106.2 \\
C++ & 24.1 & 10.5 & 63.0 & 149.3 & 101.5 \\
Java & 16.2 & 10.2 & 51.0 & 107.3 & 91.0 \\
JavaScript & 15.6 & 12.0 & 62.3 & 112.2 & 119.4 \\
TypeScript & 19.8 & 13.7 & 62.1 & 145.4 & 149.7 \\
\midrule
\textbf{Average} & \textbf{18.4} & \textbf{11.7} & \textbf{57.6} & \textbf{117.9} & \textbf{108.2} \\
\bottomrule
\end{tabular}}
\end{minipage}
\end{table*}

 \section{Evaluation methods}
\label{sec:eval}

Given the challenges in evaluating LLMs, we employ a combination of methods: functional correctness; similarity-based metrics, which offer fast, scalable evaluation across languages; and LLM-judge evaluations to assess output quality from a human-aligned perspective.

\subsection{Functional correctness}

For functional correctness, we report Pass@$1$ with $n=5$ samples~\citep{chenEvaluatingLargeLanguage2021}, measuring the probability that at least one generated sample passes all test cases: $\text{pass@}k := \mathbb{E}_{\text{Problems}} [ 1 - \binom{n-c}{k}/\binom{n}{k} ]$ where $c$ is the number of correct samples and $k=1$. Execution details are in Appendix~\ref{subsec:functional_correctness}.

\subsection{Similarity-Based evaluation}

We use two widely adopted similarity
metrics: Average Cosine Similarity and Line 0 Exact Match Rate. Average Cosine Similarity assesses
semantic equivalence across the full completion, even when syntax differs, while Line 0 Exact Match
focuses on strict precision at the start of the completion. Each metric is averaged over the $n=5$ generated completions per test case.

\textbf{Average Cosine Similarity}: We use token-based cosine similarity~\citep{zhouCodeBERTScoreEvaluatingCode2023} to measure semantic overlap between model-generated and golden completions. When tokenization fails due to unusual code constructs, we fall back to character n-grams (1-3) to ensure robust comparison.

\textbf{Line 0 Exact Match Rate}: We calculate the percentage of cases where the first line of the model-generated completion exactly matches the first line of the golden completion~\citep{dingCrossCodeEvalDiverseMultilingual2023}.

\subsection{LLM-judge evaluation}
\label{subsec:llm_judge_eval_desc}

We use an LLM judge to score each completion on two 0--5 dimensions: \textbf{relevance} to the surrounding context and \textbf{helpfulness} in advancing the task, yielding a 0--10 combined score averaged over $n=5$ completions per task. We use Gemini 2.5 Flash~\citep{Gemini25Flash} as judge because it belongs to a different provider family than all evaluated models and achieves strong code-judge accuracy on CodeJudgeBench (74.66)~\citep{CodeJudgeBench}; independent shortcut-bias audits report that Gemini 2.5 Flash exhibits smaller verdict shift rates under provenance and recency cue perturbations than GPT-4o~\citep{SilentJudge}. We blind the judge to model identity and use point-wise scoring rather than pairwise comparison to avoid pairwise response-ordering effects.

We validate the judge against 450 human-scored completions stratified by language, model, category, and score range. Three annotators independently scored each completion, with acceptable reliability (Intraclass Correlation Coefficient, ICC(3,1) = 0.76); Gemini judgments correlate with averaged human scores at Spearman $\rho = 0.81$ ($p < 0.001$), with per-language correlations ranging from $\rho = 0.74$ (C\#) to $\rho = 0.87$ (Python) and no language-level correlation falling below 0.70. Large judge--human disagreements occurred in 50 of 450 cases (11\%); full calibration details and failure analysis are in Appendix~\ref{sec:judge_failure_analysis}. Confidence intervals are estimated via 10,000 bootstrap resamples.
 \section{Experiments}
\label{sec:experiments}

\subsection{Experimental setup}
\label{subsec:experimetal_setup}

\textbf{Models}: We evaluated 9 state-of-the-art LLMs spanning six providers, three capability tiers, and both closed and open weights. Our selection includes three frontier models spanning closed and open availability (GPT-5.5~\citep{GPT55SystemCard}, Claude Opus 4.7~\citep{ClaudeOpus47}, and DeepSeek V4 Pro~\citep{DeepSeekV4Pro}); large open-weight models from Meta and Mistral (Llama 4 Maverick~\citep{Llama4Maverick} and Mistral Medium 3.5~\citep{MistralMedium35}); and mid-tier or compact models: Claude Sonnet 4.6~\citep{ClaudeSonnet46}, GPT-5.4 Mini and GPT-5.4 Nano~\citep{GPT54MiniNano}, and Qwen3.6-27B~\citep{Qwen36}. This suite covers closed frontier systems, large open-weight models, and compact/open models across six providers, including four open-weight families: DeepSeek, Meta, Mistral, and Alibaba.

\textbf{Evaluation Setup}: Following prior work~\citep{jainLiveCodeBenchHolisticContamination2024}, we set a maximum output length of 800 tokens to accommodate complex completions. Models that support temperature with reasoning enabled used a temperature of 0.2 for more deterministic completions; remaining models used their default reasoning settings. All models used nucleus sampling with top-p=1.0 to preserve the full token distribution while modulating randomness via temperature. For LLM-judge evaluation, we used Gemini 2.5 Flash~\citep{Gemini25Flash} via Vertex AI with default settings (temperature=1.0 and top-p=1.0). Models were evaluated in a zero-shot setting, each prompted using a consistent, code-only template, excluding explanations or comments. See Appendix~\ref{subsec:evaluation_prompt} for prompt details, Appendix~\ref{subsec:infrastructure} for infrastructure, and Appendix~\ref{sec:assumptions} for key assumptions.

\subsection{Results and insights}

\subsubsection{Functional correctness (Pass@1)}
\label{subsec:passrate_evaluation}

\begin{table*}[t]
  \caption{Pass@1 with $n=5$ across code completion categories.}
  \label{tab:model-performance}
  \centering
  \resizebox{\textwidth}{!}{
  \begin{tabular}{lccccccc}
    \toprule
    Model & Overall $\downarrow$ & API Usage & Code Purpose & Code2NL/NL2Code & Low Context & Pattern Matching & Syntax \\
    \midrule
    GPT-5.5 & \textbf{43.5\%} & \textbf{37.5\%} & 37.6\% & \textbf{31.5\%} & \textbf{54.2\%} & 48.8\% & \textbf{51.4\%} \\
    DeepSeek V4 Pro & 43.3\% & 34.5\% & \textbf{55.5\%} & 29.3\% & 48.0\% & \textbf{60.4\%} & 32.4\% \\
    Claude Opus 4.7 & 40.5\% & 33.8\% & 40.1\% & \textbf{31.5\%} & 46.7\% & 45.4\% & 45.7\% \\
    Llama 4 Maverick & 35.8\% & 30.6\% & 49.9\% & 22.5\% & 41.7\% & 40.6\% & 29.7\% \\
    Claude Sonnet 4.6 & 32.6\% & 31.9\% & 27.3\% & 27.1\% & 45.6\% & 27.5\% & 36.3\% \\
    Mistral Medium 3.5 & 32.1\% & 26.6\% & 45.5\% & 21.3\% & 34.8\% & 38.9\% & 25.7\% \\
    GPT-5.4 Mini & 28.2\% & 32.7\% & 16.3\% & 25.5\% & 40.3\% & 25.3\% & 29.1\% \\
    GPT-5.4 Nano & 27.1\% & 29.8\% & 15.4\% & 25.3\% & 41.1\% & 23.4\% & 27.7\% \\
    Qwen3.6-27B & 18.0\% & 18.1\% & 16.1\% & 13.8\% & 26.3\% & 22.7\% & 11.3\% \\
    \bottomrule
  \end{tabular}
  }
\end{table*}

Table~\ref{tab:model-performance} shows Pass@1 results with $n=5$ samples, averaged across the six programming languages. Language-specific Pass@1 breakdowns are provided in Table~\ref{tab:pass1-by-language}.

\textbf{Top Performers}: GPT-5.5 leads with 43.5\%, followed closely by DeepSeek V4 Pro (43.3\%) and Claude Opus 4.7 (40.5\%). No model exceeds 44\% Pass@1, confirming that \textbf{DevBench} remains challenging even for frontier reasoning models.

\textbf{Open Models}: Llama 4 Maverick (35.8\%) and Mistral Medium 3.5 (32.1\%) slot between the frontier and compact tiers. Llama outperforms Claude Sonnet 4.6 overall, while Mistral is within 0.5 percentage points of Sonnet and above the GPT-5.4 compact models, demonstrating that large open-weight models are competitive with closed mid-tier systems on hardened code completion tasks.

\textbf{Small-Size Models}: Qwen3.6-27B achieves a modest overall Pass@1 of 18.0\%, while GPT-5.4 Nano reaches 27.1\%. The roughly 25-point gap between the strongest and weakest model demonstrates clear discrimination across the capability spectrum, with a smooth gradient from frontier (40--44\%) through mid-tier open and closed models (32--36\%) to compact models (27--28\%) and the smallest open model (18\%).

\textbf{Category Patterns}: Low Context is the strongest category across models (top performers: 46--54\%), while Code2NL/NL2Code is the most challenging (leading models: 29--32\%, most below 26\%). Pattern Matching shows the largest model differentiation, with DeepSeek V4 Pro at 60.4\% and Qwen3.6-27B at 22.7\%. API Usage is relatively uniform among the frontier cluster (33--38\%) but drops sharply for smaller models (Qwen3.6-27B: 18.1\%).

\subsubsection{Similarity-Based evaluation}
\label{subsec:similarity_evaluation}

Table~\ref{tab:similarity-metrics} reports similarity metrics across languages, averaged over categories. Additional similarity results are in Appendix~\ref{subsec:full_similarity}; qualitative examples of the behaviors described below appear in Appendices~\ref{sec:model_comparison} and~\ref{sec:qualitative_examples}.

\textbf{Top Performers}: DeepSeek V4 Pro, Llama 4 Maverick, and Mistral Medium 3.5 achieve the strongest average cosine similarity scores across most languages, indicating that their completions often remain close to the golden completions even when they fail hidden assertions. DeepSeek V4 Pro retains the highest Line 0 Exact Match Rates in Python and JavaScript (60.67\% and 62.67\%), while Mistral Medium 3.5 achieves the highest cosine similarity in Python and Java.

\textbf{Metric Discrepancies}: DeepSeek V4 Pro achieves the highest similarity scores overall despite being tied with GPT-5.5 on Pass@1 (43.3\% vs 43.5\%), suggesting its completions are syntactically closer to golden completions even when they may fail hidden assertions. Conversely, Claude Opus 4.7's C\# similarity collapse (0.12 cosine) aligns with its 9.9\% Pass@1 in that language. This failure confirms a systematic language-specific gap rather than a metric artifact. This demonstrates how combining functional correctness with similarity metrics provides richer diagnostic insight than either metric alone.

\textbf{Language-Specific Challenges}: C\# shows the lowest cosine similarity for several models, while TypeScript is lowest for others. C\# is particularly challenging for Anthropic models (Opus 4.7: 0.12, Sonnet 4.6: 0.21). This language-specific variation highlights the importance of cross-language evaluation for identifying targeted improvement opportunities.

\begin{table*}[t]
  \caption{Similarity metrics across programming languages.}
  \label{tab:similarity-metrics}
  \centering
  \scriptsize
  \setlength{\tabcolsep}{4pt}
  \begin{tabular}{@{}lcccccccccccc@{}}
    \toprule
    & \multicolumn{6}{c}{Average Cosine Similarity} & \multicolumn{6}{c}{Line 0 Exact Match Rate (\%)} \\
    \cmidrule(lr){2-7} \cmidrule(lr){8-13}
    Model & Py & JS & TS & Java & C++ & C\# & Py & JS & TS & Java & C++ & C\# \\
    \midrule
    GPT-5.5 & 0.64 & 0.62 & 0.53 & 0.67 & 0.58 & 0.44 & 54.00 & 53.67 & 32.33 & 51.00 & 39.00 & 43.33 \\
    DeepSeek V4 Pro & 0.72 & 0.72 & 0.69 & 0.73 & 0.69 & 0.61 & 60.67 & 62.67 & 39.00 & 56.67 & 45.00 & 56.00 \\
    Claude Opus 4.7 & 0.63 & 0.58 & 0.56 & 0.70 & 0.61 & 0.12 & 53.67 & 49.33 & 35.33 & 54.67 & 39.00 & 14.33 \\
    Llama 4 Maverick & 0.71 & 0.70 & 0.66 & 0.70 & 0.67 & 0.67 & 53.67 & 47.67 & 26.67 & 47.33 & 37.67 & 42.33 \\
    Claude Sonnet 4.6 & 0.52 & 0.44 & 0.42 & 0.58 & 0.49 & 0.21 & 43.00 & 37.33 & 26.67 & 47.00 & 31.67 & 22.67 \\
    Mistral Med.\ 3.5 & 0.73 & 0.70 & 0.67 & 0.74 & 0.68 & 0.00 & 54.67 & 48.00 & 29.67 & 47.33 & 38.00 & 0.00 \\
    GPT-5.4 Mini & 0.47 & 0.38 & 0.30 & 0.49 & 0.46 & 0.48 & 45.33 & 38.00 & 20.67 & 43.67 & 34.00 & 41.67 \\
    GPT-5.4 Nano & 0.34 & 0.30 & 0.29 & 0.47 & 0.41 & 0.40 & 31.33 & 31.33 & 25.67 & 43.67 & 32.00 & 35.33 \\
    Qwen3.6-27B & 0.35 & 0.38 & 0.33 & 0.40 & 0.38 & 0.38 & 44.33 & 45.67 & 22.33 & 40.00 & 33.67 & 34.67 \\
    \bottomrule
  \end{tabular}
\end{table*}

\subsubsection{LLM-judge evaluation}

\begin{figure}[htbp]
    \centering
    \includegraphics[width=0.87\linewidth]{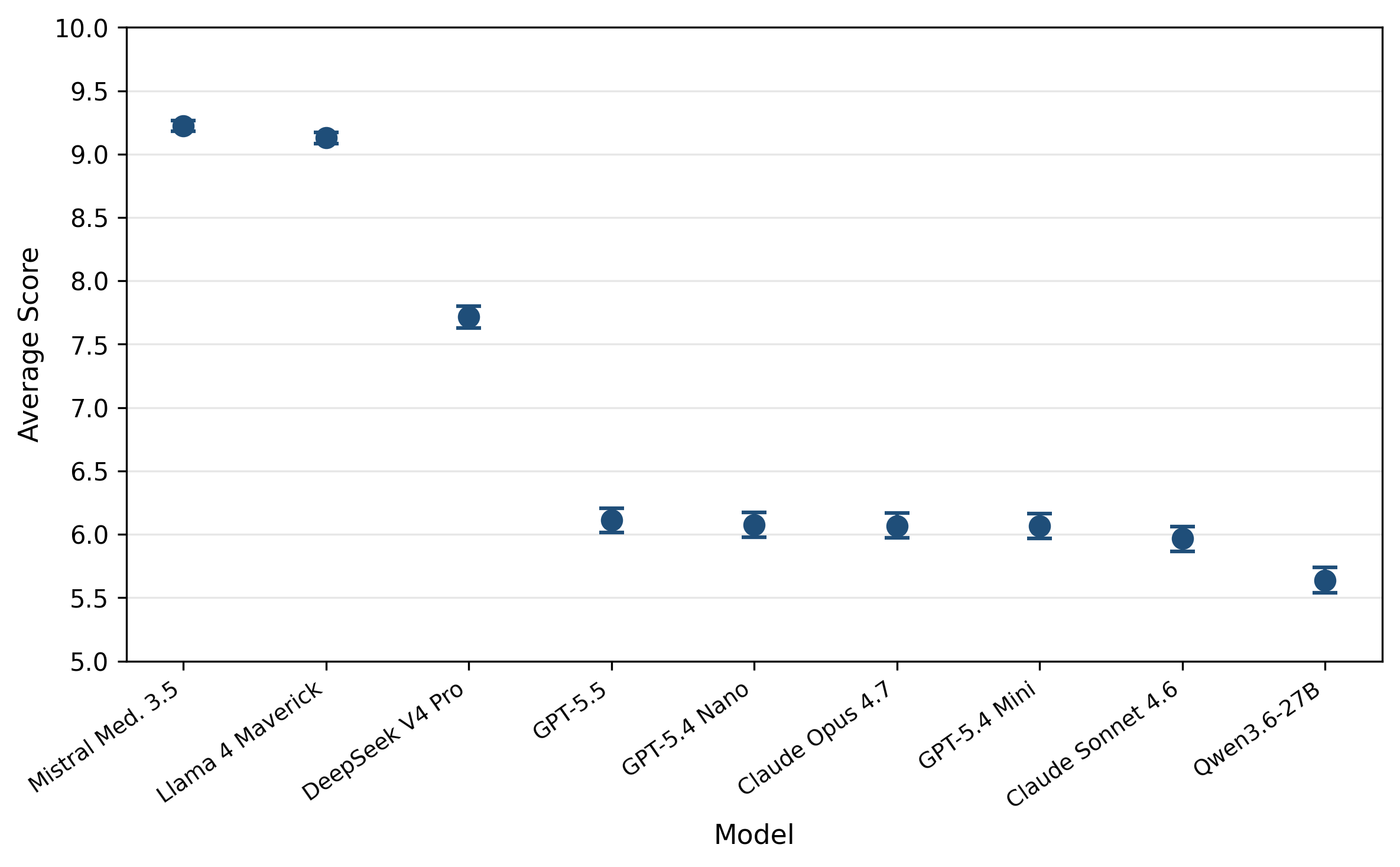}
    \caption{Overall LLM-judge evaluation scores with 95\% confidence intervals.}
    \label{fig:llm-judge-finalscore}
\end{figure}

Figure~\ref{fig:llm-judge-finalscore} presents the final LLM-judge scores with 95\% confidence intervals and Appendix~\ref{sec:llmjudge-fullresults} provides a detailed breakdown by category and language.

\textbf{LLM-judge score patterns}: The LLM-judge ranking differs substantially from Pass@1, reflecting that the judge scores relevance and helpfulness rather than executable correctness. Llama 4 Maverick and Mistral Medium 3.5 receive the highest judge scores (9.13 and 9.22 respectively) despite lower Pass@1, indicating that their completions are often contextually plausible under the relevance/helpfulness rubric, even when they fail hidden functional assertions. Among the frontier models, DeepSeek V4 Pro leads the judge at 7.71, ahead of GPT-5.5 (6.11) and Claude Opus 4.7 (6.07).

\textbf{Pass@1 vs.\ Judge Divergence}: GPT-5.5, which leads on Pass@1 (43.5\%), scores below Llama 4 Maverick and Mistral Medium 3.5 on the judge, despite those models ranking lower on Pass@1. This divergence underscores that plausible-looking completions do not necessarily satisfy hidden assertions, reinforcing why \textbf{DevBench} reports functional correctness alongside judge assessments. Because the judge scores relevance and helpfulness rather than executable correctness, divergence from Pass@1 is expected: a completion may be contextually useful but fail a hidden assertion, or it may be executable but less helpful to a developer.

\textbf{Confidence Intervals}: Most models display relatively narrow confidence intervals, indicating consistent performance across evaluation instances. The narrow intervals indicate that the judge-score differences are not driven by a small number of outlier completions.

\subsection{Diagnostic Case Study: DeepSeek V4 Pro}

\textbf{DevBench}'s multi-metric framework enables fine-grained diagnosis beyond aggregate rankings. To demonstrate its practical utility, we present a case study on DeepSeek V4 Pro, a useful diagnostic case because it is near the top on Pass@1 while exhibiting sharp category-specific divergences between similarity and functional correctness.

\textbf{Syntax vs. Semantics:} DeepSeek V4 Pro excels in Syntax Completion similarity in Table~\ref{tab:similarity-metrics-categories} (Average Cosine Similarity 0.71 vs.\ GPT-5.5's 0.67 and Line 0 Exact Match Rate 54.33\% vs.\ 53.33\%) but underperforms in functional correctness of the same category in Table~\ref{tab:model-performance} (32.4\% vs.\ 51.4\% Pass@1). This pattern indicates heavier reliance on pattern memorization than true semantic understanding. Manual review of failure cases confirms that DeepSeek V4 Pro often produces code syntactically close to the golden solution but functionally incorrect.

\textbf{Category-Level Patterns:} Based on Table~\ref{tab:similarity-metrics-categories}, the model demonstrates strong performance in Pattern Matching (0.78 vs.\ GPT-5.5's 0.56) and Code Purpose (0.76 vs.\ 0.42), where both similarity and Pass@1 confirm genuine strength. However, Code2NL/NL2Code is the model's weakest category (29.3\% Pass@1, cosine 0.56). This disparity reveals the model's tendency to excel at extending established code patterns rather than deeply understand and generate code in semantically rich tasks requiring bidirectional translation between natural language and code.

\textbf{Language-Specific Gaps:} While DeepSeek V4 Pro is the most balanced model across languages (9.4-percentage-point spread), it shows notable underperformance in TypeScript (36.1\%) relative to its strongest languages (Table~\ref{tab:pass1-by-language}). The LLM-judge scores reveal a sharper gap: 9.70 in C++ but only 5.42 in JavaScript, suggesting the model's completions vary significantly in developer-perceived quality across language ecosystems.

\textbf{Preserving Strengths:} DeepSeek V4 Pro already excels in Pattern Matching and Code Purpose Understanding, areas that should be maintained during future fine-tuning to avoid catastrophic forgetting.

These insights translate to actionable training priorities: (1) emphasize syntactic precision during fine-tuning to close the gap between similarity and functional correctness, (2) increase Code2NL/NL2Code training examples to improve semantic understanding, (3) target TypeScript and JavaScript samples in the training mix to close language-specific gaps, and (4) maintain current strengths in Pattern Matching and Code Purpose.

\section{Related Work}
\label{sec:related_work}

Existing LLM coding evaluation spans three main areas. \emph{Problem solving benchmarks} like HumanEval~\citep{chenEvaluatingLargeLanguage2021}, MBPP~\citep{austinProgramSynthesisLarge2021}, and APPS~\citep{hendrycksMeasuringCodingChallenge2021} evaluate coding problems of varying difficulty, while Concode and CoNaLa focus on natural language to code translation~\citep{iyerMappingLanguageCode2018, yinLearningMineAligned2018}. \emph{Repository-based benchmarks} evaluate code generation within existing codebases, from simple masking tasks (RepoMasterEval, ClassEval~\citep{wuRepoMasterEvalEvaluatingCode2024, duClassEvalManuallyCraftedBenchmark2023}) to inter-file reasoning (CrossCodeEval, CoderEval~\citep{dingCrossCodeEvalDiverseMultilingual2023, yuCoderEvalBenchmarkPragmatic2024}), API usage (BigCodeBench~\citep{zhuoBigCodeBenchBenchmarkingCode2025}), and agentic problem solving (SWE-Bench~\citep{jimenezSWEbenchCanLanguage2024}, SWE-Bench Pro~\citep{dengSWEBenchProCan2025}). \emph{Evolving benchmarks} like LiveCodeBench and EvoCodeBench address data contamination using recent code~\citep{jainLiveCodeBenchHolisticContamination2024, liEvoCodeBenchEvolvingCode2024}. Recent work has also scaled multi-language evaluation: M2RC-Eval~\citep{M2RCEVAL} provides repository-level code completion across 18 languages, and AutoCodeBench~\citep{AutoCodeBench} generates 3,920 contest-style problems across 20 languages. Both focus on standalone code generation rather than the FIM completion scenarios that arise in real developer workflows. In contrast, \textbf{DevBench} evaluates scenarios arising during live development, with task categories derived from telemetry analysis of real developer interactions.

\section{Conclusion}
\label{sec:conclusion}

We introduced \textbf{DevBench}, a synthetic benchmark grounded in developer telemetry, enabling fine-grained, realistic code completion evaluation across six languages and six task categories, resulting in 1,800 evaluation instances with a focus on ecological validity, contamination resistance, and interpretability. Evaluating 9 state-of-the-art models spanning six providers and four open-weight families, we observed consistent strengths in low-context pattern recognition and persistent challenges in bidirectional natural language--code translation and syntactic alignment. Our multi-pronged evaluation, combining functional correctness, similarity metrics, and LLM-judge assessments, revealed nuanced differences such as cross-language consistency and robustness across task types. By releasing the benchmark, we aim to support more accountable, targeted, and practical evaluation of code generation models. Future work could explore composite evaluation metrics and broader coverage of development activities such as refactoring, debugging, and multi-file architecture design.

\bibliographystyle{plainnat}
\bibliography{references}
\label{sec:references}

\appendix

\section*{Appendix Table of Contents}
\startcontents[appendix]
\printcontents[appendix]{l}{1}{\setcounter{tocdepth}{2}}

\section{Benchmark construction details}
\label{sec:benchmark_construction_details}

\subsection{Telemetry derivation details}
\label{sec:telemetry_derivation}

Category derivation proceeded in three stages. First, we drew a stratified, difficulty-enriched sample of 50,000 completions from the full telemetry corpus, balanced across the six target languages and weighted toward completions with low acceptance rates or high post-acceptance edit distances, since these represent scenarios where current models struggle. Sampling weights were used when estimating prevalence, while the balanced sample ensured adequate annotation coverage for lower-volume languages. Each sampled completion was tagged with telemetry signals including acceptance outcome (accepted as-is, edited, or rejected), character-level edit distance between the generated and final accepted completion, time-to-decision, and completion length. Second, three researchers independently annotated a subset of 3,000 completions (500 per language) with scenario and failure-mode labels (e.g., specialized API usage, low-context completion, pattern extension, incorrect API call, wrong control flow, missing edge case, misunderstood intent, documentation mismatch, syntax error). Inter-annotator agreement on the primary scenario/failure-mode label was substantial (Fleiss' $\kappa = 0.74$ before adjudication), indicating that the telemetry-derived categories were not artifacts of a single annotator's judgment. We clustered these annotations based on label co-occurrence, then consolidated the resulting clusters into 8--12 candidate categories per language before cross-language consolidation. Third, the research team consolidated cross-language clusters into the final six categories through iterative discussion with language specialists, merging clusters that shared underlying capability requirements and splitting those that conflated distinct skills. Categories were retained only if they met two criteria: \emph{prevalence} (the scenario or failure family accounted for $\geq$5\% of low-acceptance completions across at least four of six languages) and \emph{actionability} (the scenario could be isolated into a self-contained evaluation instance solvable from prefix and suffix alone).

The language distribution in our telemetry corpus is not uniform: Python and JavaScript together account for approximately 55\% of completions, while C++ and C\# each represent approximately 8--10\%. We chose equal representation (300 instances per language) to ensure adequate statistical power for cross-language comparison, rather than mirroring the telemetry distribution. The category distribution in the final benchmark (50 instances per language--category cell) similarly prioritizes balanced evaluation over telemetry frequency.

\textbf{DevBench} offers higher complexity and realism than prior benchmarks, with evaluation instances averaging 57.6 LOC. Importantly, \textbf{DevBench} maintains a balanced prefix-to-completion ratio: completions average 11.7 LOC, with 117.9 tokens in the prefix and 108.2 in the completion. In contrast, CrossCodeEval features long prompts (71--116 LOC) but extremely short completions (1--2 LOC). This balance makes \textbf{DevBench} more reflective of practical code-completion workflows, where both context and generated code contribute meaningfully to task complexity.

\subsection{Benchmark generation and validation details}
\label{sec:generation_validation_details}

Each generator produced 684 candidate instances, corresponding to 19 candidates for each of the 36 language-category cells, using the same structured prompts with temperature 0.7 and a 4000-token limit, yielding 2,052 total candidates. Of these, 89 failed automatic syntax or execution validation, 41 were removed by near-duplicate filtering against public benchmarks, and 122 were rejected during blinded human review, leaving 1,800 instances in the final benchmark.

Each instance consists of four components: (1) a \textbf{prefix} providing the preceding code context, (2) a \textbf{golden completion} as the expected model output, (3) a \textbf{suffix} representing subsequent code, and (4) \textbf{assertions} to validate correctness. Assertions are stored in a separate field hidden from the model for all six languages, ensuring that models cannot reverse-engineer correct completions from visible test cases. \textbf{DevBench} covers both prefix-only completions and fill-in-the-middle (FIM) cases where a suffix is provided. Completions are positioned at natural code boundaries (e.g., after operators, function calls, variable declarations) reflecting realistic developer cursor positions.

Each instance was independently reviewed by two annotators from a team of three senior researchers and engineers with expertise across all six target languages. The annotators evaluated four dimensions: (1) \emph{usefulness} (if the completion satisfies a plausible developer need), (2) \emph{realism} (if it reflects authentic coding patterns, including common inconsistencies and suboptimal but valid approaches), (3) \emph{category alignment} (if it is consistent with the intended task type), and (4) \emph{complexity authenticity} (if it captures the genuine difficulty and edge cases observed in telemetry). Inter-annotator agreement on the initial binary accept/reject decision, before adjudication, was substantial (Cohen's $\kappa = 0.82$, raw agreement 91\%); disagreements were resolved through discussion with the third annotator.

Annotators were specifically instructed to prioritize realism over idealized implementations. For example, API Usage cases were validated not only for correct library calls but also for realistic parameter handling, error conditions, and incomplete context that developers actually encounter in practice. The rejected samples, primarily due to low challenge or category mismatch, were regenerated and re-verified until they met realism standards. Common rejection reasons included: overly simplified or ``textbook-perfect'' implementations (32\% of rejections), insufficient complexity relative to telemetry-observed patterns (28\%), unrealistic examples that ignored common edge cases or error conditions (23\%), and category misalignment where the completion didn't match the intended task type (17\%). Retention rates were similar across generator families, suggesting that no single generator dominated the final benchmark.

\subsection{Generator bias and contamination analysis}
\label{sec:generator_bias_details}

\emph{Generator-family favoritism.} A benchmark generated by a single LLM may contain surface patterns that favor models from the same provider family. To test this directly, we generated 684 candidate instances per generator using three earlier-generation models from different families (GPT-4o, Claude~3.5 Sonnet, DeepSeek-V3). We fit a logistic regression predicting pass/fail for each model--instance pair, with fixed effects for evaluated model, generator family, language, category, and completion-length bin, and an indicator for whether the evaluated model and generator share a provider family; confidence intervals were estimated by bootstrap resampling over instances. The same-family coefficient was small and not statistically significant. Model rankings computed separately on each generator's instance subset were also highly correlated, with minimum pairwise Spearman $\rho = 0.91$ across the three generator subsets, and retention rates after validation, near-duplicate filtering, and human review were comparable across generators (GPT-4o: 87.3\%, Claude~3.5 Sonnet: 87.9\%, DeepSeek-V3: 88.0\%). We find no evidence of systematic generator-family favoritism, though we do not claim that all generator effects are absent~\citep{maheshwariEfficacySyntheticData2024,chenDiversitySyntheticData2024}.

\emph{Distributional overlap.} A broader concern is that LLM-generated code may reflect patterns present in public code corpora. Generator diversity alone cannot eliminate this risk, since modern code models share substantial training-data overlap. \textbf{DevBench} therefore uses generation only to instantiate telemetry-derived task structures: the categories, completion modes, cursor positions, difficulty targets, and scenario types are derived from over one billion real developer interactions, while the released code instances are synthetic and privacy-preserving. We audited all candidate instances for near-duplicate overlap against the established benchmarks in Table~\ref{tab:benchmark-complexity} using token 4-gram Jaccard similarity and AST-level structural similarity. We removed candidates with token 4-gram Jaccard similarity above 0.50 or AST structural similarity above 0.85, thresholds chosen conservatively to flag near-duplicate solutions rather than common programming idioms. Of the 2,052 candidates, 41 (2.0\%) exceeded these thresholds and were removed. Manual inspection of all 41 flagged pairs and a random sample of 100 unflagged pairs confirmed that the thresholds primarily captured near-duplicate task structure rather than generic language constructs or common API boilerplate. AST structural similarity is computed over normalized AST node-type sequences after removing identifiers, literals, and formatting.

\subsection{Within-prompt diversity analysis}
\label{sec:within_prompt_diversity}

Because each of the 36 language--category prompts generates multiple instances, a concern is whether instances from the same prompt are insufficiently diverse. To test this, we computed average pairwise cosine similarity (using TF-IDF token vectors) among all instances sharing the same prompt. The overall mean pairwise similarity is 0.094, corresponding to an average cosine distance of 0.906, indicating that instances generated from the same prompt are highly distinct under this token-based measure. Per-language means range from 0.060 (Python, most diverse) to 0.116 (C++, least diverse, reflecting greater boilerplate from includes and namespaces). Per-category means range from 0.077 (API Usage) to 0.117 (Pattern Matching, where shared structural patterns are expected by design). Even the least diverse language--category cell (C\# Pattern Matching, 0.159) remains below 16\% similarity. These results indicate that prompt specialization does not produce homogeneous instances.

\subsection{Benchmark generation prompts}
\label{subsec:benchmark_generation_prompts}

Each language--category cell has a dedicated generation prompt encoding the telemetry-derived task definition, output schema, hidden-assertion format, difficulty requirements, and rejection criteria. All three generator models received the same prompt for a given cell without modification, enabling the cross-generator bias analysis in Section~\ref{subsec:generator_bias_analysis}. Below we reproduce the C++ API Usage prompt as a representative example; the complete set of 36 prompts is released with the benchmark repository.

\begin{center}
\textbf{C++: API Usage Prompts}
\end{center}

\begin{lstlisting}[style=pythonstyle]
API_USAGE_SYSTEM_PROMPT = """
You are an expert C++ benchmark designer creating realistic
code-completion evaluation instances for large language models.

Your task is to generate one high-quality C++ API Usage instance
that reflects telemetry-observed developer completion scenarios.
The instance should test whether a model can correctly use a
common but error-prone C++ API under realistic context
constraints, including
parameter ordering, resource management, error handling, type
conversions, API-specific conventions, and edge cases.

The instance must be a code-completion task, not a standalone
programming problem. The evaluated model will see only:
  - prefix: code before the cursor
  - suffix: optional code after the cursor for fill-in-the-middle settings

The evaluated model will NOT see:
  - golden_completion
  - assertions
  - LLM_justification

Return a single valid JSON object with the following fields:
  - id: unique identifier
  - testsource: "devbench-api-usage"
  - language: "cpp"
  - prefix: code before the completion point
  - suffix: code after the completion point (may be empty)
  - golden_completion: the minimal correct code at the cursor
  - assertions: hidden executable test code used only for
    evaluation; must not duplicate or leak the golden completion
  - LLM_justification: why this is a realistic and challenging
    API Usage task

Requirements:

1. Realistic API usage.
   Use APIs from the C++ standard library and common development
   contexts, reflecting telemetry-observed completion scenarios.
   Representative domains include:
   - Standard algorithms: sort with custom comparators, transform
     (unary/binary), partition, nth_element, clamp, rotate, search
     with searcher objects, copy_backward, adjacent_find
   - Smart pointers and memory management: unique_ptr ownership
     transfer, shared_ptr/make_shared semantics, weak_ptr::lock()
     promotion, enable_shared_from_this, allocator patterns
   - Threading and concurrency: async/future, condition_variable
     with predicates, shared_mutex reader-writer locks, scoped_lock
     for deadlock-free multi-mutex acquisition, semaphore patterns
   - Container semantics: map insert vs operator[], reserve vs
     resize, emplace return values, set extract and node handles
   - Move semantics and forwarding: std::move, std::forward for
     perfect forwarding, std::exchange in move constructors
   - Tuple utilities: std::tie with std::ignore, std::apply for
     tuple-to-argument expansion, std::tuple_cat
   - Numeric algorithms: accumulate vs reduce, iota, inner_product
     with custom binary operations
   - String and text processing: substr boundary handling, stoi
     with index output parameter, string_view lifetime semantics,
     regex_replace
   - Filesystem: create_directories with error_code overloads,
     path decomposition and lexical normalization
   - C++17 vocabulary types: variant with std::visit, optional
     emplace vs assignment, any_cast value-vs-pointer overloads
   - Low-level type support: from_chars/to_chars, std::bitset,
     std::launder with placement new, aligned_storage
   - Chrono timing: steady_clock vs system_clock, duration_cast
   - Callable abstractions: std::invoke with member pointers
   The API call should be embedded in a plausible function, class,
   or workflow, not presented as isolated trivia.

2. Difficulty.
   The completion should require resolving at least two contextual
   constraints from the prefix/suffix (type compatibility,
   ownership/lifetime, error-path behavior, parameter ordering,
   boundary conditions, or consistency with a helper method). The
   task should not be solvable by copying a nearby line. Avoid
   textbook-perfect toy examples; include realistic engineering
   context such as fallback behavior or resource cleanup.

3. Completion structure.
   The golden_completion must contain only the code at the cursor.
   The suffix must not duplicate the golden_completion. The prefix
   and suffix together must make the completion inferable but not
   reveal the answer. The combined prefix + golden_completion +
   suffix + assertions must compile and execute successfully.

4. Hidden assertions.
   Assertions must be stored only in the "assertions" field; do
   not place hidden tests in the prefix or suffix. Assertions must
   validate functional behavior, not just syntax. Include at least
   one edge case not explicitly described in comments. Assertions
   must not hard-code or leak the golden completion.

5. Rejection criteria. Do not generate instances that are:
   ambiguous (multiple equally valid completions), dependent on
   unavailable external services, near-duplicated from public
   benchmarks, solvable by a single obvious keyword, dominated
   by boilerplate rather than API reasoning, invalid C++, or
   overly simplified textbook implementations.

Output a single-line JSON object with all newlines and quotes
escaped for JSONL parsing.
"""

API_USAGE_USER_PROMPT = """
Generate one C++ API Usage evaluation instance. Choose a domain
from the representative list in the system prompt (standard
algorithms, smart pointers, threading, containers, move
semantics, tuples, numerics, strings, filesystem, C++17
vocabulary types, low-level type support, chrono, or callable
abstractions).

CRITICAL JSON FORMATTING REQUIREMENTS:
1. Your response MUST be a syntactically valid JSON object
2. PROPERLY ESCAPE all special characters in strings:
   - Use \\" for double quotes inside strings
   - Use \\n for newlines, \\t for tabs, \\\\ for backslashes
3. The entire JSON object must be on a SINGLE LINE
4. DO NOT use markdown code blocks in your response

Required JSON fields:
- id: unique numeric identifier
- testsource: "devbench-api-usage"
- language: "cpp"
- prefix: code before the completion point (establishes context)
- suffix: code after the completion point; must be DIFFERENT from
  the golden_completion; may contain execution code but NOT
  hidden assertions
- golden_completion: the minimal correct code at the cursor that
  maintains consistency with prefix/suffix
- assertions: hidden executable test code used ONLY for
  evaluation; the model under test will never see this field;
  must validate functional behavior and include at least one
  edge case not spelled out in comments
- LLM_justification: why this is a realistic, challenging task

CRITICAL: HIDDEN ASSERTION REQUIREMENTS:
1. All functional test code must go in the "assertions" field
2. The "assertions" field must NOT be empty
3. Do NOT place hidden tests in the prefix or suffix
4. Assertions must not hard-code or leak the golden completion
5. The combined prefix + golden_completion + suffix + assertions
   must compile and execute successfully
6. Include at least one edge case assertion

COMPLETION STRUCTURE REQUIREMENTS:
1. The golden_completion must contain ONLY the code at the cursor
2. The suffix must NOT duplicate the golden_completion
3. The prefix and suffix must make the completion inferable but
   not reveal the answer
4. The completion should require resolving at least two
   contextual constraints from prefix/suffix

PREFIX LENGTH REQUIREMENTS:
1. The prefix MUST be at least 15-25 lines of code
2. Provide sufficient context and setup code
3. Include helper functions, type definitions, or related code
4. The prefix should demonstrate an incomplete implementation

CODE STRUCTURE REQUIREMENTS:
1. All code must be within proper C++ scope boundaries
2. Do not place executable statements or assertions at global
   scope; includes, type aliases, constants, class/struct
   definitions, and helper function definitions are allowed
3. All code blocks must have matching braces
4. Include only headers that are actually used
5. The code must be fully executable C++

INDENTATION REQUIREMENTS:
1. All code sections must maintain consistent indentation
2. The golden_completion must match the prefix indentation level
3. The suffix must maintain the same indentation context

Format your response as a single-line JSON object:
{"id": "1", "testsource": "devbench-api-usage", "language":
"cpp", "prefix": "...", "suffix": "...",
"golden_completion": "...", "assertions": "...",
"LLM_justification": "..."}

VALIDATION CHECKLIST:
1. Is your response a single, valid JSON object?
2. Are all special characters properly escaped?
3. Are assertions in the "assertions" field (NOT in the suffix)?
4. Does prefix + golden_completion + suffix + assertions compile?
5. Does the golden_completion resolve multiple context constraints?
6. Is the task non-trivial and not solvable by copying a line?
7. Are all required JSON fields present, with non-empty prefix,
   golden_completion, assertions, and LLM_justification? The
   suffix may be empty for prefix-only instances.
8. Does at least one assertion test an edge case?
"""
\end{lstlisting}

\section{Evaluation methodology details}
\label{sec:eval_methodology_details}

\subsection{Infrastructure}
\label{subsec:infrastructure}

Our benchmark generation and evaluation workloads were distributed across cloud-based model APIs and local computing resources. Model API calls were orchestrated from a standard laptop (11th Gen Intel i7-1165G7 @ 2.80GHz with 16GB RAM) running Python 3.10. For benchmark generation, using three generator APIs (OpenAI GPT-4o, Anthropic Claude~3.5 Sonnet, and DeepSeek-V3) to create synthetic evaluation instances required approximately 6--12 hours of total wall-clock time across all generators and languages, depending on API latency and excluding human review time. Each individual model evaluation on the complete benchmark required approximately 1.5-3 hours of wall-clock time, also dependent on the API latency. The execution component of our evaluation pipeline, which verifies functional correctness, was executed on the same laptop and required approximately 15 minutes per model (details in Appendix~\ref{subsec:functional_correctness}). 

\subsection{Functional Correctness Evaluation}
\label{subsec:functional_correctness}

Our functional correctness evaluation methodology implements robust, secure, and reproducible execution environments across all six programming languages. Each evaluation instance consists of four components: a context prefix, a golden completion (or model-generated completion during evaluation), a context suffix, and assertion statements that verify correctness. The execution pipeline combines these components into complete, executable programs with language-specific safeguards and dependency management.

\textbf{Python Execution Environment.} Python evaluation instances run in controlled subprocesses with 30-second timeouts to prevent infinite loops. We automatically insert \texttt{matplotlib} non-interactive backend configuration to prevent \texttt{plt.show()} calls from blocking execution, and handle environment variables securely to provide necessary API access while maintaining isolation. When dependency-related errors occur, our system automatically attempts to install missing packages using \texttt{pip} before retrying execution. Each evaluation instance runs in its own isolated environment to prevent cross-contamination between tests.

\textbf{Java Execution Environment.} Java evaluation uses adaptive compilation strategies based on code complexity. Simple test cases without external dependencies use direct \texttt{javac} compilation and execution with assertions enabled via the \texttt{-ea} flag. Complex cases requiring external libraries (Apache Commons, Jackson, Guava, etc.) automatically utilize Gradle build management with Maven Central dependency resolution. Our system detects package declarations and import statements to determine the appropriate compilation strategy, ensuring both basic and enterprise-level Java code can be properly evaluated.

\textbf{JavaScript and TypeScript Execution.} JavaScript evaluation uses Node.js execution with automatic \texttt{npm} package installation for missing dependencies. We configure execution environments with proper PATH resolution to ensure consistent Node.js and npm access across different system configurations. TypeScript evaluation adds a compilation step using \texttt{tsc} before JavaScript execution, with automatic installation of TypeScript compiler and type definitions (\texttt{@types} packages) as needed. Both environments support up to five retry attempts for dependency resolution to handle multiple missing packages.

\textbf{C\# Execution Environment.} C\# evaluation employs \texttt{dotnet run} for basic console applications and full MSBuild project compilation for complex scenarios requiring NuGet packages. Our system automatically detects namespace declarations and external dependencies (Entity Framework, Newtonsoft.Json, Azure SDKs, etc.) to generate appropriate \texttt{.csproj} files with package references. The execution environment targets .NET 6.0 for broad compatibility while supporting modern C\# language features.

\textbf{C++ Execution Environment.} C++ evaluation uses multiple compiler detection (\texttt{g++}, \texttt{clang++}) with comprehensive library path resolution for external dependencies. Our system automatically detects common libraries (OpenSSL, Boost, OpenCV, Eigen, etc.) from include statements and configures appropriate compiler flags, include paths, and library linking. We support both Homebrew and system-installed libraries across macOS and Linux platforms, with automatic detection of architecture-specific paths (Apple Silicon vs Intel).

\textbf{Cross-Language Safeguards.} All execution environments implement consistent safety measures: isolated temporary directories with automatic cleanup, configurable timeouts (30-60 seconds based on language compilation requirements), comprehensive error handling with detailed diagnostic reporting, and proper resource management to prevent system interference. Dependencies are installed locally within test directories rather than globally to maintain system isolation.

This multi-language execution infrastructure allows us to comprehensively evaluate functional correctness across diverse programming paradigms while maintaining the security and reproducibility essential for reliable benchmarking. The entire evaluation pipeline generates both human-readable reports and structured JSON output for detailed analysis of model performance across languages and categories.

\subsection{Evaluation prompt}
\label{subsec:evaluation_prompt}

For our model evaluation process, we implemented a carefully designed prompt template focused on precise code completion tasks. After initial experimentation revealed that different prompt formats could significantly impact model performance due to formatting issues, we selected a structured instruction-based approach that addresses common failure modes. Our code repository contains the full evaluation prompt.

The selected prompt format provides clear examples demonstrating proper replacement behavior in various scenarios, explicitly instructing models to maintain correct indentation and avoid duplicating existing code structures. By standardizing the input format with clear \texttt{\#TODO: You Code Here markers} and providing explicit instructions against common mistakes, we created a more level evaluation environment that better isolates models' code understanding capabilities from prompt interpretation abilities.

This evaluation prompt design aligns with real-world code completion scenarios where maintaining contextual formatting is essential for functional correctness, ensuring our benchmark more accurately reflects models' practical utility in development environments. Performance differences observed between models using this standardized prompt more reliably indicate their intrinsic code completion capabilities rather than their ability to navigate ambiguous or unstructured prompting patterns.

\subsection{LLM-judge calibration and failure analysis}
\label{sec:judge_failure_analysis}

Judge calibration proceeded in two stages. First, we iteratively tuned the judge prompt on telemetry acceptance signals from 10,000 completions (accepted as-is, rejected, edited after acceptance), selecting relevance and helpfulness criteria that best separated accepted-as-is, edited, and rejected completions. Low scores correspond to completions that are syntactically plausible but irrelevant to the surrounding context, fail to advance the intended task, omit required edge-case handling, or contradict the prefix/suffix constraints. Second, we validated on a held-out, stratified set of 450 completions (75 per language, sampled across models, categories, and score ranges) with three experienced annotators independently scoring each completion on a 0--10 rubric. The 450-completion validation is intended as a held-out audit of rank-order alignment rather than as a replacement for exhaustive human evaluation; the primary calibration signal comes from the 10,000-completion telemetry tuning set.

For each model, we compute average scores by programming language and evaluation scenario. We then aggregate completions within each language to obtain an overall average and a 95\% confidence interval, estimated via 10,000 bootstrap resamples. Finally, we report the overall average score across all languages and completions with its corresponding confidence interval.

Of the 450 validation completions, 50 (11\%) exhibited disagreements exceeding 2 points between the averaged human score and the Gemini 2.5 Flash judge score. We manually categorized these into two failure directions.

\textbf{Judge scores too high} (31 of 50 cases). The most common pattern (19 cases) involved completions that were syntactically well-formed and contextually plausible but contained subtle logical errors (such as off-by-one boundary conditions, incorrect operator precedence, or wrong variable references) that human annotators identified from the surrounding context but the judge, which cannot execute code, did not penalize. The remaining 12 cases involved over-generation: the model produced additional code beyond what the task required (e.g., unrequested helper functions or extra validation branches) that the judge rewarded for apparent helpfulness while human annotators penalized for deviating from the expected completion scope.

\textbf{Judge scores too low} (19 of 50 cases). The majority (13 cases) occurred in Code2NL/NL2Code instances, where the judge weighted documentation completeness more heavily than human annotators, penalizing terse but accurate comments that omitted optional tags (e.g., missing \texttt{@param} entries in an otherwise correct Javadoc block). The remaining 6 cases involved completions using unconventional but functionally correct idioms (such as ternary chains instead of if/else blocks, or compact list comprehensions instead of explicit loops) that the judge rated as less readable despite human annotators scoring them as acceptable.

These patterns suggest that the judge's primary vulnerability is an inability to verify functional correctness (scoring too high on plausible-looking but buggy code) and a tendency toward stricter documentation standards than human annotators apply in practice (scoring too low on terse but accurate completions). Both failure modes are consistent with the judge evaluating relevance and helpfulness without execution feedback rather than executable correctness, which is why we report Pass@1 as a complementary metric.

\section{Full experimental results}
\label{sec:full_experimental_results}

\subsection{Full similarity metrics by category}
\label{subsec:full_similarity}

We provide the complete similarity-based results across all evaluated models in Table~\ref{tab:similarity-metrics-categories}. This expanded view offers a more comprehensive comparison of model performance across different categories and similarity dimensions.

\begin{table*}[t]
  \caption{Similarity metrics across task categories. Most models use reasoning/thinking capabilities; Llama 4 Maverick and Qwen3.6-27B are non-reasoning models.}
  \label{tab:similarity-metrics-categories}
  \centering
  \resizebox{\textwidth}{!}{
  \begin{tabular}{lcccccccccccc}
    \toprule
    & \multicolumn{6}{c}{Average Cosine Similarity} & \multicolumn{6}{c}{Line 0 Exact Match Rate (\%)} \\
    \cmidrule(lr){2-7} \cmidrule(lr){8-13}
    Model & API & Code2NL & Purpose & Low & Pattern & Syntax & API & Code2NL & Purpose & Low & Pattern & Syntax \\
    & Usage & NL2Code & Underst. & Context & Matching & Compl. & Usage & NL2Code & Underst. & Context & Matching & Compl. \\
    \midrule
    GPT-5.5 & 0.67 & 0.51 & 0.42 & 0.65 & 0.56 & 0.67 & 46.67 & 48.00 & 39.00 & 42.00 & 44.33 & 53.33 \\
    DeepSeek V4 Pro & 0.67 & 0.56 & 0.76 & 0.67 & 0.78 & 0.71 & 46.33 & 53.00 & 64.33 & 44.00 & 58.00 & 54.33 \\
    Claude Opus 4.7 & 0.60 & 0.47 & 0.48 & 0.52 & 0.50 & 0.61 & 39.67 & 40.67 & 44.33 & 36.33 & 38.33 & 47.00 \\
    Llama 4 Maverick & 0.63 & 0.56 & 0.74 & 0.67 & 0.79 & 0.72 & 40.00 & 45.67 & 51.33 & 29.67 & 46.67 & 42.00 \\
    Claude Sonnet 4.6 & 0.59 & 0.44 & 0.29 & 0.48 & 0.32 & 0.53 & 38.00 & 43.00 & 31.00 & 30.33 & 26.33 & 39.67 \\
    Mistral Medium 3.5 & 0.54 & 0.42 & 0.63 & 0.60 & 0.70 & 0.61 & 29.67 & 35.00 & 43.33 & 30.33 & 43.67 & 35.67 \\
    GPT-5.4 Mini & 0.63 & 0.52 & 0.23 & 0.47 & 0.33 & 0.40 & 43.67 & 46.00 & 27.00 & 33.00 & 36.33 & 37.33 \\
    GPT-5.4 Nano & 0.55 & 0.48 & 0.15 & 0.40 & 0.22 & 0.40 & 43.67 & 48.00 & 19.67 & 27.00 & 24.33 & 36.67 \\
    Qwen3.6-27B & 0.44 & 0.40 & 0.28 & 0.36 & 0.42 & 0.30 & 41.00 & 45.33 & 36.00 & 28.67 & 37.00 & 32.67 \\
    \bottomrule
  \end{tabular}
  }
\end{table*}

\subsection{Pass@1 Performance by Programming Language}

Table~\ref{tab:pass1-by-language} provides a detailed breakdown of Pass@1 performance across all six programming languages for each evaluated model. Python remains the easiest language for most models (GPT-5.5: 53.6\%, Claude Opus 4.7: 54.1\%), while the hardest language varies by model: C\# is hardest for 4 models, C++ for 2, and TypeScript for 3. DeepSeek V4 Pro is the most balanced across languages, with only a 9.4-percentage-point spread (36.1--45.5\%), compared to Claude Opus 4.7's 44.2-point spread (9.9--54.1\%). One model shows a severe C\# weakness: Claude Opus 4.7 at 9.9\%, despite scoring competitively in all other languages. This reinforces that cross-language evaluation reveals capability gaps invisible in Python-only benchmarks.

\begin{table*}[t]
  \caption{Pass@1 rates by programming language using $n=5$ samples (sorted by overall Pass@1 $\downarrow$). Most models use reasoning/thinking capabilities; Llama 4 Maverick and Qwen3.6-27B are non-reasoning models.}
  \label{tab:pass1-by-language}
  \centering
  \resizebox{\textwidth}{!}{
  \begin{tabular}{lcccccc}
    \toprule
    Model & Python & JavaScript & TypeScript & Java & C++ & C\# \\
    \midrule
    GPT-5.5 & 53.6\% & \textbf{49.2\%} & 38.9\% & 47.2\% & 38.3\% & 33.7\% \\
    DeepSeek V4 Pro & 45.2\% & 45.5\% & 36.1\% & 44.5\% & \textbf{45.0\%} & \textbf{43.7\%} \\
    Claude Opus 4.7 & \textbf{54.1\%} & 47.5\% & \textbf{42.5\%} & \textbf{47.8\%} & 41.5\% & 9.9\% \\
    Llama 4 Maverick & 41.6\% & 30.9\% & 30.5\% & 37.5\% & 35.9\% & 38.6\% \\
    Claude Sonnet 4.6 & 44.5\% & 36.7\% & 32.0\% & 36.0\% & 27.9\% & 18.6\% \\
    Mistral Medium 3.5 & 44.6\% & 35.5\% & 33.5\% & 38.9\% & 40.1\% & 0.0\% \\
    GPT-5.4 Mini & 31.5\% & 34.2\% & 21.6\% & 28.8\% & 23.2\% & 29.9\% \\
    GPT-5.4 Nano & 30.5\% & 29.3\% & 23.8\% & 27.5\% & 23.7\% & 28.1\% \\
    Qwen3.6-27B & 22.4\% & 20.3\% & 14.7\% & 19.6\% & 13.5\% & 17.7\% \\
    \bottomrule
  \end{tabular}
  }
\end{table*}

\subsection{Category correlation analysis}
\label{sec:category_correlations}

Table~\ref{tab:category-correlations} reports pairwise Spearman correlations of Pass@1 rates across all 54 model--language combinations (9 evaluated models $\times$ 6 languages).

\begin{table}[h]
\caption{Pairwise Spearman correlations of Pass@1 across categories ($n = 54$ model--language pairs).}
\label{tab:category-correlations}
\centering
\small
\begin{tabular}{lcccccc}
\toprule
 & API & Code2NL & Purpose & Low Ctx & Pattern & Syntax \\
\midrule
API Usage    & 1.00 &      &      &      &      &      \\
Code2NL      & 0.58 & 1.00 &      &      &      &      \\
Purpose      & 0.16 & 0.38 & 1.00 &      &      &      \\
Low Context  & 0.36 & 0.48 & 0.54 & 1.00 &      &      \\
Pattern      & 0.17 & 0.37 & 0.74 & 0.58 & 1.00 &      \\
Syntax       & 0.36 & 0.48 & 0.36 & 0.60 & 0.46 & 1.00 \\
\bottomrule
\end{tabular}
\end{table}

These correlations are computed over aggregate category Pass@1 scores, so they should be interpreted as an empirical diagnostic of category relationships rather than as a formal latent-factor model of code-generation ability.

Several pairs show low correlation: API Usage vs.\ Code Purpose Understanding ($\rho = 0.16$) and API Usage vs.\ Pattern Matching ($\rho = 0.17$), reflecting that library-specific knowledge draws on different capabilities than semantic reasoning or pattern extension over existing code. Conversely, Pattern Matching and Code Purpose Understanding correlate more highly ($\rho = 0.74$), consistent with both requiring semantic reasoning; however, these categories still diverge diagnostically for individual models. This analysis is intended to characterize empirical category relationships, not to prove that the categories are mutually exclusive; real developer tasks naturally blend capabilities, and correlated categories still provide diagnostic value when their rankings diverge for a specific model.

\subsection{Detailed LLM-judge experimental results}
\label{sec:llmjudge-fullresults}
\begin{figure}[htbp]
    \centering
    \includegraphics[width=0.82\textwidth]{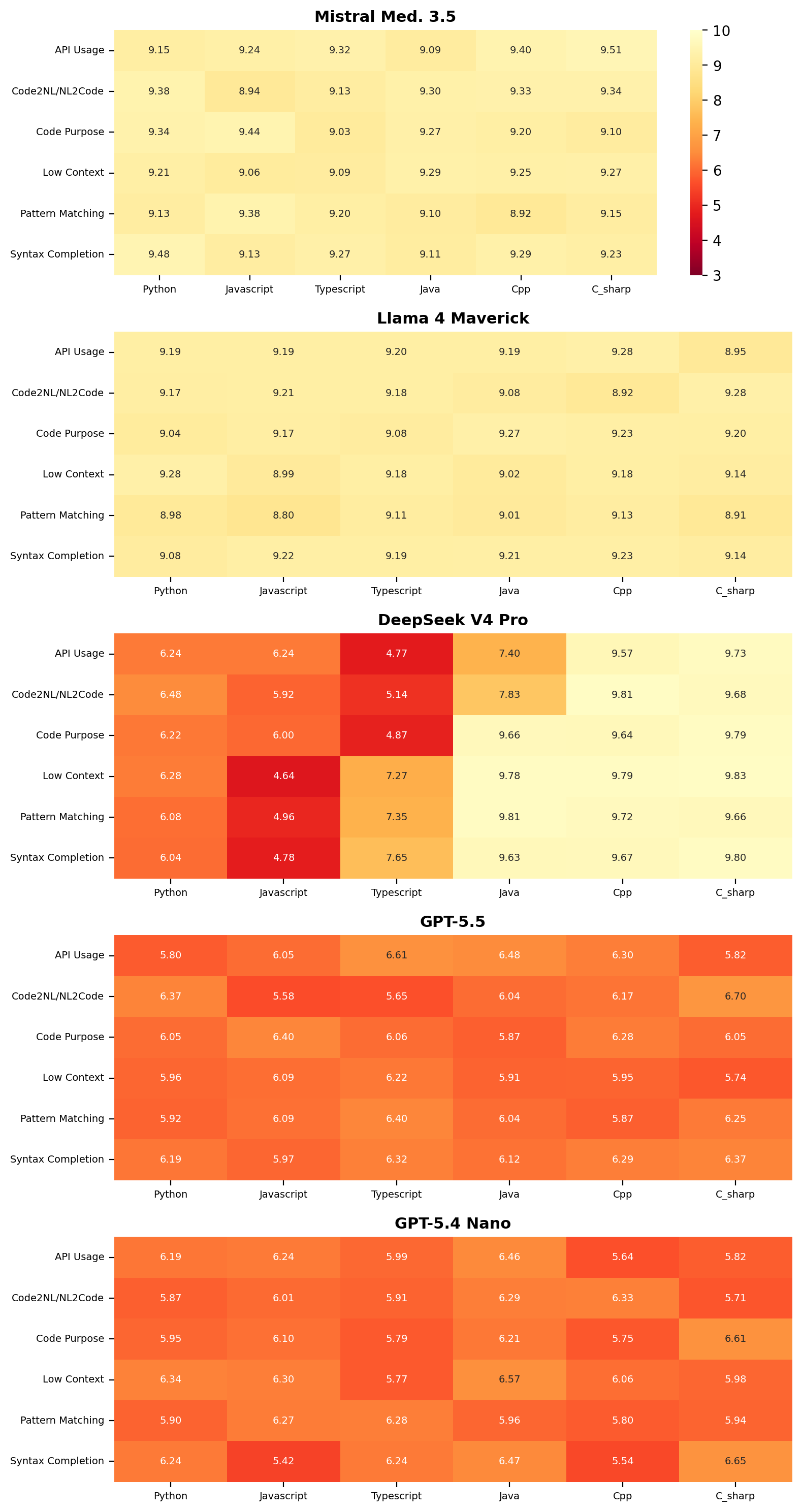}
    \caption{LLM-judge scores by category and language for the top five models (sorted by overall judge score).}
    \label{fig:llm-judge-fullscores}
\end{figure}

\begin{figure}[htbp]
    \centering
    \includegraphics[width=0.9\textwidth]{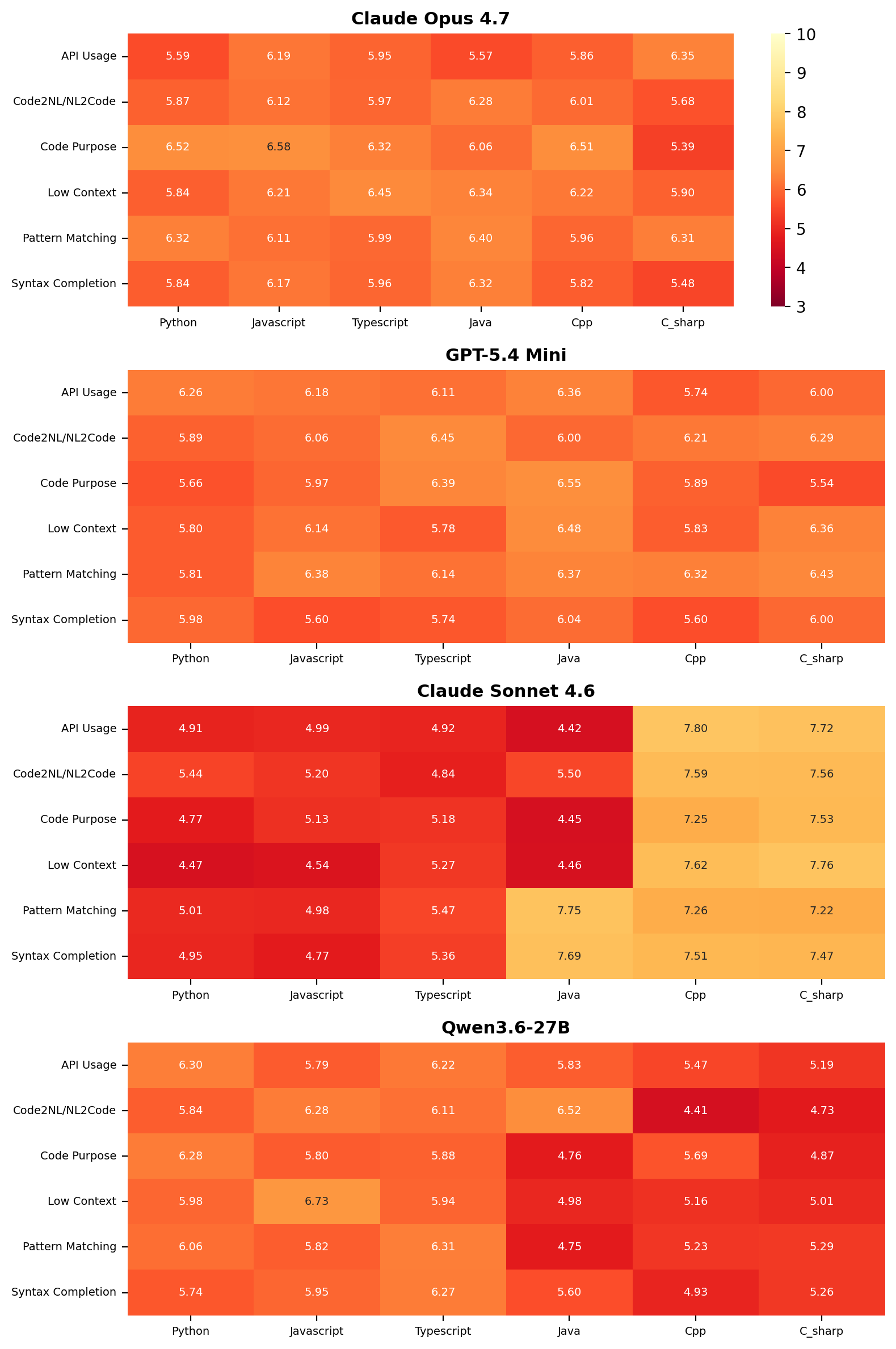}
    \caption{LLM-judge scores by category and language for the remaining four models (continued from Figure~\ref{fig:llm-judge-fullscores}).}
    \label{fig:llm-judge-fullscores-2}
\end{figure}

Figures~\ref{fig:llm-judge-fullscores} and~\ref{fig:llm-judge-fullscores-2} show the breakdown of LLM-judge scores by category and language.

To better understand the relative performance of different models across programming languages, Table~\ref{tab:language-scores-llm-judge} presents the LLM-judge scores with 95\% confidence intervals for each model-language pair.

\begin{table*}[t]
  \caption{Programming language LLM-judge scores of different LLMs with 95\% confidence intervals (sorted by overall judge score $\downarrow$).}
  \label{tab:language-scores-llm-judge}
  \centering
  \resizebox{\textwidth}{!}{
  \begin{tabular}{lcccccc}
    \toprule
    Model & C++ & C\# & Java & JavaScript & Python & TypeScript \\
    \midrule
    Mistral Medium 3.5 & 9.23 (9.13-9.33) & 9.26 (9.16-9.36) & \textbf{9.19} (9.09-9.29) & \textbf{9.20} (9.09-9.30) & \textbf{9.28} (9.18-9.38) & \textbf{9.17} (9.06-9.28) \\
    Llama 4 Maverick & 9.16 (9.06-9.26) & 9.10 (9.00-9.21) & 9.13 (9.02-9.24) & 9.10 (8.98-9.21) & 9.12 (9.01-9.23) & 9.16 (9.05-9.26) \\
    DeepSeek V4 Pro & \textbf{9.70} (9.63-9.76) & \textbf{9.75} (9.69-9.80) & 9.02 (8.87-9.16) & 5.42 (5.18-5.67) & 6.22 (5.99-6.47) & 6.17 (5.93-6.41) \\
    GPT-5.5 & 6.14 (5.90-6.38) & 6.16 (5.92-6.39) & 6.08 (5.83-6.32) & 6.03 (5.79-6.27) & 6.05 (5.80-6.29) & 6.21 (5.97-6.45) \\
    GPT-5.4 Nano & 5.85 (5.61-6.10) & 6.12 (5.88-6.36) & 6.33 (6.09-6.56) & 6.06 (5.82-6.29) & 6.08 (5.85-6.32) & 6.00 (5.76-6.24) \\
    Claude Opus 4.7 & 6.06 (5.82-6.30) & 5.85 (5.61-6.10) & 6.16 (5.92-6.40) & 6.23 (5.99-6.47) & 6.00 (5.75-6.24) & 6.11 (5.87-6.35) \\
    GPT-5.4 Mini & 5.93 (5.69-6.16) & 6.10 (5.86-6.34) & 6.30 (6.06-6.54) & 6.05 (5.81-6.30) & 5.90 (5.65-6.14) & 6.10 (5.86-6.34) \\
    Claude Sonnet 4.6 & 7.51 (7.29-7.72) & 7.54 (7.33-7.76) & 5.71 (5.47-5.95) & 4.93 (4.68-5.17) & 4.93 (4.68-5.17) & 5.17 (4.93-5.41) \\
    Qwen3.6-27B & 5.15 (4.91-5.38) & 5.06 (4.81-5.30) & 5.41 (5.16-5.65) & 6.06 (5.82-6.30) & 6.03 (5.78-6.28) & 6.12 (5.88-6.36) \\
    \bottomrule
  \end{tabular}
  }
\end{table*}

\clearpage

\section{Qualitative analysis}
\label{sec:qualitative_analysis}

\subsection{Further category examples}
\label{sec:category_examples}

\subsubsection{API Usage}

This example tests knowledge of C\# 8.0's Index and Range operators introduced for expressive array slicing. The prefix declares an integer array and a comment directing the model to use \texttt{\^{}} (index from end) and \texttt{..} (range) operators. The golden completion demonstrates six distinct uses: \texttt{data[\^{}1]} for the last element, \texttt{data[\^{}2]} for second-to-last, \texttt{data[2..6]} for a middle slice, \texttt{data[\^{}3..]} for the last three elements, \texttt{data[..2]} for the first two, and \texttt{data[\^{}4..\^{}1]} for a reversed-index range. The suffix validates each slice's content, while hidden assertions test edge cases including \texttt{data[\^{}8]} (first element via full-length index), \texttt{data[..]} (full copy), and \texttt{data[3..3]} (empty range). Models must understand both the \texttt{\^{}} from-end semantics and the half-open \texttt{..} range semantics simultaneously.

\begin{center}
\textbf{Example 1: C\# API Usage \#43}
\end{center}

\begin{lstlisting}[style=csharpstyle]
using System;

namespace Bench
{
    public class Program
    {
        static void Main()
        {
            int[] data = { 10, 20, 30, 40, 50, 60, 70, 80 };

            // Use Index (^) and Range (..) operators to extract slices
\end{lstlisting}

\begin{lstlisting}[style=golden, language={[Sharp]C}]
            int last = data[^1];
            int secondLast = data[^2];
            int[] middle = data[2..6];
            int[] lastThree = data[^3..];
            int[] firstTwo = data[..2];
            int[] reversed = data[^4..^1];
\end{lstlisting}

\begin{lstlisting}[style=suffix, language={[Sharp]C}]

            if (!(last == 80)) throw new Exception("last");
            if (!(secondLast == 70)) throw new Exception("secondLast");
            if (!(middle.Length == 4 && middle[0] == 30 && middle[3] == 60)) throw new Exception("middle");
            if (!(lastThree.Length == 3 && lastThree[0] == 60)) throw new Exception("lastThree");
            if (!(firstTwo.Length == 2 && firstTwo[1] == 20)) throw new Exception("firstTwo");
            if (!(reversed.Length == 3 && reversed[0] == 50 && reversed[2] == 70)) throw new Exception("reversed");
            Console.WriteLine("All slices correct");
        }
    }
}
\end{lstlisting}

\subsubsection{Code2NL/NL2Code}

This example evaluates the NL-to-Code direction: the model must implement a thread-safe lazy singleton from a detailed Javadoc specification. The prefix provides the complete specification (double-checked locking with a \texttt{volatile} field, fast-path null check, synchronized block, and second null check) along with the field declarations and constructor. The golden completion implements \texttt{getInstance()} following the spec exactly: outer null check, \texttt{synchronized (C2NTask19.class)}, inner null check, and instance creation. Hidden assertions verify singleton identity and use reflection to confirm the \texttt{instance} field is declared \texttt{volatile}, a subtle but critical requirement that prevents instruction reordering on multicore processors. Models that skip the outer null check (unnecessary synchronization), omit the inner null check (race condition), or forget \texttt{volatile} (memory visibility bug) all fail.

\begin{center}
\textbf{Example 2: Java Code2NL/NL2Code \#19}
\end{center}

\begin{lstlisting}[style=javastyle]
public class C2NTask19 {

    /**
     * Thread-safe lazy singleton using double-checked locking.
     * The instance field MUST be declared volatile to prevent
     * instruction reordering. The getInstance() method must:
     *   1. Check if instance is null without synchronization (fast path)
     *   2. If null, synchronize on the class object
     *   3. Check again inside the synchronized block (double-check)
     *   4. Create instance only if still null
     * The constructor is private and sets initialized=true.
     */
    private static volatile C2NTask19 instance;
    private final boolean initialized;

    private C2NTask19() {
        this.initialized = true;
    }

    public boolean isInitialized() { return initialized; }

    public static C2NTask19 getInstance() {
\end{lstlisting}

\begin{lstlisting}[style=golden, language=Java]
        if (instance == null) {
            synchronized (C2NTask19.class) {
                if (instance == null) {
                    instance = new C2NTask19();
                }
            }
        }
        return instance;
\end{lstlisting}

\begin{lstlisting}[style=suffix, language=Java]
    }

    public static void main(String[] args) {
        C2NTask19 s1 = C2NTask19.getInstance();
        C2NTask19 s2 = C2NTask19.getInstance();
        assert s1 == s2 : "must be same instance";
        assert s1.isInitialized() : "must be initialized";
        System.out.println("Task 19 PASS");
    }
}
\end{lstlisting}

\subsubsection{Low Context}

This example tests deep JavaScript knowledge from minimal context. The prefix contains only a comment and a function signature (3 lines total), yet the model must know that JavaScript's \texttt{Proxy} object with a \texttt{has} trap intercepts the \texttt{in} operator, and that \texttt{Number()} coercion is needed to handle string property keys. The golden completion creates a Proxy with \texttt{has} and \texttt{get} traps: \texttt{has} checks numeric membership in a range, \texttt{get} returns the bounds. Hidden assertions verify negative ranges, boundary values, and string-number coercion (\texttt{'3' in r2} must return \texttt{true}). At only 13 total lines, this task demonstrates that Low Context completion requires recognizing language-specific patterns from minimal cues.

\begin{center}
\textbf{Example 3: JavaScript Low Context \#16}
\end{center}

\begin{lstlisting}[style=javascriptstyle]
// Creates a range object where 'in' operator checks numeric membership
// and .min/.max properties return bounds
function createRange(min, max) {
\end{lstlisting}

\begin{lstlisting}[style=golden, language=JavaScript]
    return new Proxy({}, {
        has(_, p) { const n = Number(p); return !isNaN(n) && n >= min && n <= max; },
        get(_, p) { if (p === 'min') return min; if (p === 'max') return max; }
    });
\end{lstlisting}

\begin{lstlisting}[style=suffix, language=JavaScript]
}
const r = createRange(1, 10);
const assert = require('assert');
assert.strictEqual(5 in r, true);
assert.strictEqual(15 in r, false);
assert.strictEqual('abc' in r, false);
assert.strictEqual(r.min, 1);
assert.strictEqual(r.max, 10);
console.log('Task 16 passed');
\end{lstlisting}

\subsubsection{Syntax Completion}

This example tests C++ template syntax with partial specialization. The prefix defines a primary \texttt{element\_type<T>} trait that returns \texttt{T} by default, followed by a comment requesting the \texttt{std::vector} specialization. The golden completion writes \texttt{template <typename T, typename Alloc> struct element\_type<std::vector<T, Alloc>{}>} with nested angle brackets, syntactically demanding because the model must correctly nest \texttt{<T, Alloc>} inside \texttt{<std::vector<...>{}>}. The suffix provides the \texttt{std::basic\_string} specialization as a pattern reference and uses \texttt{static\_assert} with \texttt{std::is\_same\_v} for compile-time verification. No runtime assertions are needed; the compiler itself validates correctness.

\begin{center}
\textbf{Example 4: C++ Syntax Completion \#41}
\end{center}

\begin{lstlisting}[style=cppstyle]
#include <iostream>
#include <vector>
#include <string>
#include <type_traits>
#include <cassert>

// element_type<T> is a type trait that extracts the element type from
// container types. For non-container types, it returns the type itself.
//
// Supported containers:
//   std::vector<T, Alloc>        -> T
//   std::basic_string<C, T, A>   -> C  (the character type)
//   everything else              -> T itself (identity)
//
// This is useful for writing generic code that needs to know what a
// container holds without depending on a specific container type.

// Primary template: non-container types map to themselves
template <typename T>
struct element_type {
    using type = T;  // fallback for non-containers
};

// Partial specialization for std::vector
\end{lstlisting}

\begin{lstlisting}[style=golden, language=C++]
template <typename T, typename Alloc>
struct element_type<std::vector<T, Alloc>> {
    using type = T;
};
\end{lstlisting}

\begin{lstlisting}[style=suffix, language=C++]

// Partial specialization for std::basic_string
template <typename CharT, typename Traits, typename Alloc>
struct element_type<std::basic_string<CharT, Traits, Alloc>> {
    using type = CharT;
};

template <typename T>
using element_type_t = typename element_type<T>::type;

int main() {
    static_assert(std::is_same_v<element_type_t<int>, int>);
    static_assert(std::is_same_v<element_type_t<std::vector<double>>, double>);
    static_assert(std::is_same_v<element_type_t<std::vector<std::string>>, std::string>);
    static_assert(std::is_same_v<element_type_t<std::string>, char>);

    std::cout << "All static_asserts passed!" << std::endl;
    return 0;
}
\end{lstlisting}

\subsubsection{Pattern Matching}

This example tests pattern recognition with a subtle trap. The prefix contains two generator functions (\texttt{chunkArray} and \texttt{chunkString}) sharing identical structure: \texttt{for (let i = 0; i < X.length; i += size) \{ yield X.slice(i, i + size); \}}. The golden completion implements \texttt{slidingWindow}, structurally similar but with two critical differences: \texttt{i += step} instead of \texttt{i += size}, and \texttt{i <= arr.length - size} instead of \texttt{i < arr.length}. A model that blindly copies the chunking pattern produces non-overlapping windows instead of sliding ones. Hidden assertions test window-equals-array, step-larger-than-1, and empty-when-array-smaller-than-window edge cases.

\begin{center}
\textbf{Example 5: JavaScript Pattern Matching \#47}
\end{center}

\begin{lstlisting}[style=javascriptstyle]
function* chunkArray(arr, size) {
    // Non-overlapping chunks
    for (let i = 0; i < arr.length; i += size) {
        yield arr.slice(i, i + size);
    }
}

function* chunkString(str, size) {
    // Same pattern for strings
    for (let i = 0; i < str.length; i += size) {
        yield str.slice(i, i + size);
    }
}

\end{lstlisting}

\begin{lstlisting}[style=golden, language=JavaScript]
function* slidingWindow(arr, size, step = 1) {
    // Overlapping chunks: advance by step (not size)
    for (let i = 0; i <= arr.length - size; i += step) {
        yield arr.slice(i, i + size);
    }
}
\end{lstlisting}

\begin{lstlisting}[style=suffix, language=JavaScript]
const assert = require('assert');
// Non-overlapping chunks
assert.deepStrictEqual([...chunkArray([1,2,3,4,5], 2)], [[1,2],[3,4],[5]]);
assert.deepStrictEqual([...chunkString('abcde', 2)], ['ab','cd','e']);

// Sliding window: overlapping with step=1 by default
const windows = [...slidingWindow([1,2,3,4], 3)];
assert.deepStrictEqual(windows, [[1,2,3],[2,3,4]]);
// With step=2
const windows2 = [...slidingWindow([1,2,3,4,5], 3, 2)];
assert.deepStrictEqual(windows2, [[1,2,3],[3,4,5]]);
console.log('Task 47 suffix passed');
\end{lstlisting}

\subsubsection{Code Purpose Understanding}
\label{subsec:code_purpose_understanding_example}

This example evaluates whether a model can infer business logic constraints from a state machine. The prefix defines a \texttt{Subscription} class with \texttt{VALID\_TRANSITIONS} (pending$\to$active, active$\to$paused/cancelled, etc.) and a \texttt{\_transition} helper that validates transitions and logs them to \texttt{auditLog}. The golden completion implements \texttt{activate(ts)} by calling \texttt{this.\_transition('active', ts)} and recording \texttt{this.activatedAt = ts}, reusing the existing helper rather than setting state directly. Hidden assertions verify that double-activation from the \texttt{active} state throws an error, and that activating a \texttt{cancelled} subscription throws. This tests whether the model reads the state machine constraints and delegates to the validation logic already present in the codebase.

\begin{center}
\textbf{Example 6: JavaScript Code Purpose Understanding \#1}
\end{center}

\begin{lstlisting}[style=javascriptstyle]
const assert = require('assert');

class Subscription {
    static VALID_TRANSITIONS = {
        pending: ['active'],
        active: ['paused', 'cancelled'],
        paused: ['active', 'cancelled'],
        cancelled: []
    };

    constructor(userId, plan) {
        this.userId = userId;
        this.plan = plan;
        this.state = 'pending';
        this.auditLog = [];
        this.activatedAt = null;
        this.pausedAt = null;
        this.totalPausedMs = 0;
    }

    _transition(newState, ts) {
        const allowed = Subscription.VALID_TRANSITIONS[this.state];
        if (!allowed || !allowed.includes(newState)) {
            throw new Error(`Cannot transition from ${this.state} to ${newState}`);
        }
        const old = this.state;
        this.state = newState;
        this.auditLog.push({ from: old, to: newState, ts });
    }
\end{lstlisting}

\begin{lstlisting}[style=golden, language=JavaScript]
    activate(ts) {
        this._transition('active', ts);
        this.activatedAt = ts;
    }
\end{lstlisting}

\begin{lstlisting}[style=suffix, language=JavaScript]
    pause(ts) {
        this._transition('paused', ts);
        this.pausedAt = ts;
    }
}

const s1 = new Subscription('u1', 'pro');
s1.activate(1000);
assert.strictEqual(s1.state, 'active');
assert.strictEqual(s1.activatedAt, 1000);
assert.strictEqual(s1.auditLog.length, 1);
assert.deepStrictEqual(s1.auditLog[0], { from: 'pending', to: 'active', ts: 1000 });
\end{lstlisting}

\subsection{Illustrative model comparison}
\label{sec:model_comparison}

\subsubsection{C++ Syntax Completion: GPT-5.5 vs.\ DeepSeek V4 Pro}

The prefix defines \texttt{tuple\_sig\_impl} using C++17 index sequences and fold expressions. The completion must implement the function body AND define a separate \texttt{tuple\_sig} wrapper that the suffix's \texttt{main()} calls. GPT-5.5 correctly generates both: a fold expression body using comma-separated \texttt{type\_tag} calls, then closes the function and defines the forwarding \texttt{tuple\_sig} template. DeepSeek V4 Pro produces a single elegant fold expression \texttt{return "(" + ( ... + (...) ) + ")"} that correctly implements the immediate function logic but never closes the function body or defines \texttt{tuple\_sig}. This causes an ``undeclared identifier'' error when \texttt{main()} calls \texttt{tuple\_sig(t1)}. The failure pattern is consistent across all 8 non-GPT-5.5 models: each completes only the immediate function body without recognizing that the suffix requires additional definitions. This demonstrates that Syntax Completion requires structural awareness, understanding what identifiers and definitions the surrounding code depends on, not just syntactic knowledge of fold expressions. The example explains the 19-percentage-point gap between GPT-5.5 and DeepSeek V4 Pro on Syntax Completion.

\begin{center}
\textbf{Example 7: C++ Syntax Completion \#6}
\end{center}

\begin{lstlisting}[style=cppstyle]
#include <type_traits>
#include <tuple>
#include <utility>
#include <string>
#include <cassert>

// Compile-time type name via constexpr if + template specialization.
// Returns a string tag for the type.
template<typename T>
std::string type_tag() {
    if constexpr (std::is_same_v<T, int>) return "i";
    else if constexpr (std::is_same_v<T, double>) return "d";
    else if constexpr (std::is_same_v<T, std::string>) return "s";
    else return "?";
}

// Build a "signature" string from a tuple's types, e.g. "(i,d,s)"
// Uses index_sequence to expand the tuple types at compile time.
template<typename Tuple, std::size_t... Is>
std::string tuple_sig_impl(std::index_sequence<Is...>) {
\end{lstlisting}

\begin{lstlisting}[style=golden, language=C++]
    std::string result = "(";
    int idx = 0;
    ((result += (idx++ > 0 ? "," : "") +
      type_tag<std::tuple_element_t<Is, Tuple>>()), ...);
    return result + ")";
}

template<typename... Ts>
std::string tuple_sig(const std::tuple<Ts...>&) {
    return tuple_sig_impl<std::tuple<Ts...>>(
        std::index_sequence_for<Ts...>{});
\end{lstlisting}

\renewcommand{\modelname}{GPT-5.5}
\begin{lstlisting}[style=model, language=C++]
    std::string sig = "(";
    ((sig += (Is == 0 ? "" : ","), sig += type_tag<std::tuple_element_t<Is, Tuple>>()), ...);
    sig += ")";
    return sig;
}

template<typename... Ts>
std::string tuple_sig(const std::tuple<Ts...>&) {
    return tuple_sig_impl<std::tuple<Ts...>>(std::index_sequence_for<Ts...>{});
\end{lstlisting}

\renewcommand{\modelname}{DeepSeek V4 Pro}
\begin{lstlisting}[style=model, language=C++]
    return "(" + ( ... + (type_tag<std::tuple_element_t<Is, Tuple>>() + (Is + 1 == sizeof...(Is) ? "" : ",")) ) + ")";
\end{lstlisting}

\begin{lstlisting}[style=suffix, language=C++]
}

int main() {
    auto t1 = std::make_tuple(1, 2.0, std::string("hi"));
    assert(tuple_sig(t1) == "(i,d,s)");

    auto t2 = std::make_tuple(42);
    assert(tuple_sig(t2) == "(i)");

    auto t3 = std::make_tuple(1.0, 2.0);
    assert(tuple_sig(t3) == "(d,d)");
    return 0;
}
\end{lstlisting}

\subsubsection{TypeScript Code Purpose: Claude Opus 4.7 vs.\ DeepSeek V4 Pro}

The prefix defines a \texttt{BankingService} class with a \texttt{TransferAudit} interface specifying \texttt{status: 'completed' | 'failed'} and a separate \texttt{reason?: string} field. The golden completion implements \texttt{transfer()} to return \texttt{\{ status: 'failed', reason: 'daily\_limit\_exceeded' \}} when validation fails. DeepSeek V4 Pro (and all other models except Opus) instead writes \texttt{status: 'daily\_limit\_exceeded'}, embedding the error type directly into the status field rather than using the interface's two-field pattern. DeepSeek also omits pushing failed transfers to the audit log. Only Claude Opus 4.7 correctly interprets the TypeScript interface structure, producing \texttt{status: 'failed'} with the reason in the dedicated field. This demonstrates that Code Purpose Understanding requires reading type definitions as behavioral specifications, not just syntactic constraints.

\begin{center}
\textbf{Example 8: TypeScript Code Purpose Understanding \#31}
\end{center}

\begin{lstlisting}[style=typescriptstyle]
interface Account {
    id: string;
    balance: number;
    dailyTransferred: number;
    lastTransferDay: number;
}

interface TransferAudit {
    from: string;
    to: string;
    gross: number;
    fee: number;
    net: number;
    day: number;
    status: string;
    reason?: string;
}

class BankingService {
    accounts: Map<string, Account> = new Map();
    dailyLimit: number;
    feeRate: number;        // fraction, e.g. 0.02 = 2%
    minFee: number;         // minimum fee per transfer
    auditLog: TransferAudit[] = [];

    constructor(dailyLimit: number, feeRate: number, minFee: number) {
        this.dailyLimit = dailyLimit;
        this.feeRate = feeRate;
        this.minFee = minFee;
    }

    addAccount(id: string, balance: number): void {
        this.accounts.set(id, { id, balance, dailyTransferred: 0, lastTransferDay: -1 });
    }

    // Transfer `amount` from `fromId` to `toId` on `day`.
    // Fee = max(amount * feeRate, minFee), deducted from SENDER in addition to amount.
    // Total debit = amount + fee.
    // Check order:
    //   1. Both accounts exist, else 'account_not_found'.
    //   2. Reset dailyTransferred if day differs from lastTransferDay.
    //   3. dailyTransferred + amount > dailyLimit? 'daily_limit_exceeded'.
    //   4. Sender balance < amount + fee? 'insufficient_funds'.
    //   5. Success: debit sender (amount+fee), credit receiver (amount only),
    //      update dailyTransferred, log audit with gross/fee/net.
\end{lstlisting}

\begin{lstlisting}[style=golden, language=TypeScript]
    transfer(fromId: string, toId: string, amount: number, day: number): TransferAudit {
        const from = this.accounts.get(fromId);
        const to = this.accounts.get(toId);
        if (!from || !to) {
            const audit: TransferAudit = { from: fromId, to: toId, gross: amount, fee: 0, net: 0, day, status: 'failed', reason: 'account_not_found' };
            this.auditLog.push(audit);
            return audit;
        }
        if (from.lastTransferDay !== day) {
            from.dailyTransferred = 0;
            from.lastTransferDay = day;
        }
        if (from.dailyTransferred + amount > this.dailyLimit) {
            const audit: TransferAudit = { from: fromId, to: toId, gross: amount, fee: 0, net: 0, day, status: 'failed', reason: 'daily_limit_exceeded' };
            this.auditLog.push(audit);
            return audit;
        }
        const fee = Math.max(Math.round(amount * this.feeRate * 100) / 100, this.minFee);
        const totalDebit = amount + fee;
        if (from.balance < totalDebit) {
            const audit: TransferAudit = { from: fromId, to: toId, gross: amount, fee, net: 0, day, status: 'failed', reason: 'insufficient_funds' };
            this.auditLog.push(audit);
            return audit;
        }
        from.balance -= totalDebit;
        to.balance += amount;
        from.dailyTransferred += amount;
        const audit: TransferAudit = { from: fromId, to: toId, gross: amount, fee, net: amount, day, status: 'success' };
        this.auditLog.push(audit);
        return audit;
    }
\end{lstlisting}

\renewcommand{\modelname}{Claude Opus 4.7}
\begin{lstlisting}[style=model, language=TypeScript]
    transfer(fromId: string, toId: string, amount: number, day: number): TransferAudit {
        const from = this.accounts.get(fromId);
        const to = this.accounts.get(toId);
        if (!from || !to) {
            const audit: TransferAudit = {
                from: fromId, to: toId, gross: amount, fee: 0, net: 0, day,
                status: 'failed', reason: 'account_not_found'
            };
            this.auditLog.push(audit);
            return audit;
        }

        if (from.lastTransferDay !== day) {
            from.dailyTransferred = 0;
            from.lastTransferDay = day;
        }

        if (from.dailyTransferred + amount > this.dailyLimit) {
            const audit: TransferAudit = {
                from: fromId, to: toId, gross: amount, fee: 0, net: 0, day,
                status: 'failed', reason: 'daily_limit_exceeded'
            };
            this.auditLog.push(audit);
            return audit;
        }

        const fee = Math.max(amount * this.feeRate, this.minFee);
        if (from.balance < amount + fee) {
            const audit: TransferAudit = {
                from: fromId, to: toId, gross: amount, fee, net: amount, day,
                status: 'failed', reason: 'insufficient_funds'
            };
            this.auditLog.push(audit);
            return audit;
        }

        from.balance -= amount + fee;
        to.balance += amount;
        from.dailyTransferred += amount;

        const audit: TransferAudit = {
            from: fromId, to: toId, gross: amount, fee, net: amount, day,
            status: 'success'
        };
        this.auditLog.push(audit);
        return audit;
    }
\end{lstlisting}

\renewcommand{\modelname}{DeepSeek V4 Pro}
\begin{lstlisting}[style=model, language=TypeScript]
    transfer(fromId: string, toId: string, amount: number, day: number): TransferAudit {
        const fromAccount = this.accounts.get(fromId);
        const toAccount = this.accounts.get(toId);
        if (!fromAccount || !toAccount) {
            return {
                from: fromId, to: toId, gross: amount, fee: 0, net: 0, day,
                status: 'account_not_found',
                reason: 'One or both accounts not found'
            };
        }
        if (fromAccount.lastTransferDay !== day) {
            fromAccount.dailyTransferred = 0;
            fromAccount.lastTransferDay = day;
        }
        if (fromAccount.dailyTransferred + amount > this.dailyLimit) {
            return {
                from: fromId, to: toId, gross: amount, fee: 0, net: 0, day,
                status: 'daily_limit_exceeded',
                reason: 'Daily limit exceeded'
            };
        }
        const fee = Math.max(amount * this.feeRate, this.minFee);
        const totalDebit = amount + fee;
        if (fromAccount.balance < totalDebit) {
            return {
                from: fromId, to: toId, gross: amount, fee, net: 0, day,
                status: 'insufficient_funds',
                reason: 'Insufficient funds'
            };
        }
        fromAccount.balance -= totalDebit;
        toAccount.balance += amount;
        fromAccount.dailyTransferred += amount;
        const audit: TransferAudit = {
            from: fromId, to: toId, gross: amount, fee, net: amount, day,
            status: 'success'
        };
        this.auditLog.push(audit);
        return audit;
    }
\end{lstlisting}

\begin{lstlisting}[style=suffix, language=TypeScript]
}

const assert = require('assert');

const bank = new BankingService(1000, 0.02, 5);
bank.addAccount('A', 500);
bank.addAccount('B', 200);

// Normal transfer: amount=100, fee=max(2, 5)=5, total debit=105
const r1 = bank.transfer('A', 'B', 100, 1);
assert.strictEqual(r1.status, 'success');
assert.strictEqual(r1.fee, 5);
assert.strictEqual(bank.accounts.get('A')!.balance, 395);
assert.strictEqual(bank.accounts.get('B')!.balance, 300);

// Large transfer: amount=300, fee=max(6, 5)=6, total=306
const r2 = bank.transfer('A', 'B', 300, 1);
assert.strictEqual(r2.fee, 6);
assert.strictEqual(bank.accounts.get('A')!.balance, 89);

console.log('Task 31 suffix passed');
\end{lstlisting}

\subsection{Qualitative examples}
\label{sec:qualitative_examples}

\subsubsection{Failed execution, dissimilar completion}

The prefix provides a \texttt{Result<T>} union type and a \texttt{Step<A,B>} type alias. Claude Opus 4.7 generates function overloads, a TypeScript best practice for type-safe pipelines, with signatures for 1, 2, and 3 steps, plus a general implementation signature. However, every overload requires at least one step argument. The hidden assertion \texttt{pipeline(1)} invokes the function with zero steps, which matches none of the overloads and produces TS2554 (``Expected 2--4 arguments, but got 1''). DeepSeek V4 Pro uses a simpler rest parameter \texttt{...steps: Step<any, any>[]} that accepts any number of steps including zero. This demonstrates a paradox where more sophisticated TypeScript knowledge (generating type-safe overloads) produces code that over-constrains the API relative to simpler, more permissive implementations.

\begin{center}
\textbf{Example 9: TypeScript Low Context \#29}
\end{center}

\begin{lstlisting}[style=typescriptstyle]
type Result<T> = { ok: true; value: T } | { ok: false; error: string };
type Step<A, B> = (a: A) => Result<B>;
\end{lstlisting}

\begin{lstlisting}[style=golden, language=TypeScript]
function pipeline<T>(input: T, ...steps: Step<any, any>[]): Result<any> {
    let current: Result<any> = { ok: true, value: input };
    for (const step of steps) {
        if (!current.ok) return current;
        current = step(current.value);
    }
    return current;
}
\end{lstlisting}

\renewcommand{\modelname}{Claude Opus 4.7}
\begin{lstlisting}[style=model, language=TypeScript]
function pipeline<A, B>(input: A, s1: Step<A, B>): Result<B>;
function pipeline<A, B, C>(input: A, s1: Step<A, B>, s2: Step<B, C>): Result<C>;
function pipeline<A, B, C, D>(input: A, s1: Step<A, B>, s2: Step<B, C>, s3: Step<C, D>): Result<D>;
function pipeline(input: any, ...steps: Step<any, any>[]): Result<any> {
    let current: Result<any> = { ok: true, value: input };
    for (const step of steps) {
        if (!current.ok) return current;
        current = step(current.value);
    }
    return current;
}
\end{lstlisting}

\renewcommand{\modelname}{DeepSeek V4 Pro}
\begin{lstlisting}[style=model, language=TypeScript]
function pipeline<T>(init: T, ...steps: Step<any, any>[]): Result<any> {
    let result: Result<any> = { ok: true, value: init };
    for (const step of steps) {
        if (!result.ok) break;
        result = step(result.value);
    }
    return result;
}
\end{lstlisting}

\begin{lstlisting}[style=suffix, language=TypeScript]
import { strict as assert } from "assert";
const double: Step<number, number> = n => ({ ok: true, value: n * 2 });
const failIfBig: Step<number, number> = n => n > 100 ? { ok: false, error: "too big" } : { ok: true, value: n };
const stringify: Step<number, string> = n => ({ ok: true, value: String(n) });
assert.deepEqual(pipeline(5, double, stringify), { ok: true, value: "10" });
assert.deepEqual(pipeline(60, double, failIfBig), { ok: false, error: "too big" });
\end{lstlisting}

\subsubsection{Failed execution, similar completion}

The prefix establishes helper functions \texttt{shallowEqual} (with epsilon for floats) and \texttt{arrayEqual} (order-independent comparison via sorting). Both GPT-5.5 and DeepSeek V4 Pro implement \texttt{deepEqual} with identical structure: null checks, type checks, recursive object comparison with sorted keys. The critical difference is a single guard clause. GPT-5.5 writes \texttt{if (Array.isArray(a) || Array.isArray(b))} followed by a nested check \texttt{if (!Array.isArray(a) || !Array.isArray(b)) return false}, which handles the mixed case where one argument is an array and the other is a plain object. DeepSeek writes \texttt{if (Array.isArray(a) \&\& Array.isArray(b))} which only enters the array branch when both are arrays, letting mixed array/object pairs fall through to object comparison. The hidden assertion \texttt{deepEqual([], \{\}) === false} exploits this: an empty array and empty object both have zero enumerable keys, so the object-comparison path returns \texttt{true}. This one missing guard (\texttt{||} vs \texttt{\&\&}) is the entire difference between 5/5 and 1/5. The example demonstrates how structurally similar completions can contain subtle logical errors that only surface under specific edge cases.

\begin{center}
\textbf{Example 10: JavaScript Pattern Matching \#44}
\end{center}

\begin{lstlisting}[style=javascriptstyle]
function shallowEqual(a, b) {
    if (typeof a !== typeof b) return false;
    if (typeof a === 'number') return Math.abs(a - b) < 1e-9;
    return a === b;
}

function arrayEqual(a, b) {
    // Pattern: arrays are compared as SORTED SETS (order-independent)
    if (a.length !== b.length) return false;
    const sa = [...a].sort((x, y) => JSON.stringify(x) < JSON.stringify(y) ? -1 : 1);
    const sb = [...b].sort((x, y) => JSON.stringify(x) < JSON.stringify(y) ? -1 : 1);
    return sa.every((v, i) => deepEqual(v, sb[i]));
}

\end{lstlisting}

\begin{lstlisting}[style=golden, language=JavaScript]
function deepEqual(a, b) {
    if (a === b) return true;
    if (a == null || b == null) return a === b;
    if (typeof a === 'number' && typeof b === 'number') return Math.abs(a - b) < 1e-9;
    if (typeof a !== typeof b) return false;
    if (Array.isArray(a) !== Array.isArray(b)) return false;
    if (Array.isArray(a)) return arrayEqual(a, b);
    if (typeof a !== 'object') return false;
    const ka = Object.keys(a).sort(), kb = Object.keys(b).sort();
    if (ka.length !== kb.length) return false;
    return ka.every((k, i) => k === kb[i] && deepEqual(a[k], b[k]));
}
\end{lstlisting}

\renewcommand{\modelname}{GPT-5.5}
\begin{lstlisting}[style=model, language=JavaScript]
function deepEqual(a, b) {
    if (a === null || b === null) return a === b;
    if (typeof a !== typeof b) return false;
    if (typeof a !== 'object') return shallowEqual(a, b);
    if (Array.isArray(a) || Array.isArray(b)) {
        if (!Array.isArray(a) || !Array.isArray(b)) return false;
        return arrayEqual(a, b);
    }

    const ka = Object.keys(a).sort();
    const kb = Object.keys(b).sort();
    if (ka.length !== kb.length) return false;

    return ka.every((key, i) => key === kb[i] && deepEqual(a[key], b[key]));
}

\end{lstlisting}

\renewcommand{\modelname}{DeepSeek V4 Pro}
\begin{lstlisting}[style=model, language=JavaScript]
function deepEqual(a, b) {
    if (a === b) return true;
    if (a == null || b == null) return a === b; // handles null/undefined correctly
    if (Array.isArray(a) && Array.isArray(b)) {
        return arrayEqual(a, b);
    }
    if (typeof a === 'object' && typeof b === 'object') {
        const keysA = Object.keys(a);
        const keysB = Object.keys(b);
        if (keysA.length !== keysB.length) return false;
        return keysA.every(key => deepEqual(a[key], b[key]));
    }
    return shallowEqual(a, b);
}
\end{lstlisting}

\begin{lstlisting}[style=suffix, language=JavaScript]
const assert = require('assert');
// Numbers with epsilon
assert.strictEqual(deepEqual(0.1 + 0.2, 0.3), true);
// Arrays as sets (order independent)
assert.strictEqual(deepEqual([1, 2, 3], [3, 1, 2]), true);
assert.strictEqual(deepEqual([1, 2], [2, 3]), false);
// Nested objects
assert.strictEqual(deepEqual({a: 1}, {a: 1}), true);
assert.strictEqual(deepEqual({a: 1}, {a: 2}), false);
// Null handling
assert.strictEqual(deepEqual(null, null), true);
assert.strictEqual(deepEqual(null, 0), false);
console.log('Task 44 suffix passed');
\end{lstlisting}

\subsubsection{Passed execution, dissimilar completion}

The prefix specifies a \texttt{slugify} function that must lowercase input, replace non-alphanumeric characters with hyphens, collapse multiple hyphens, strip leading/trailing hyphens, AND insert hyphens at camelCase boundaries and letter-digit boundaries. The golden completion implements this as a single-pass \texttt{StringBuilder} loop: it iterates character-by-character, checking the previous character against the current to detect camelCase transitions (\texttt{isLowerCase(p) \&\& isUpperCase(ch)}) and letter-digit boundaries, building the result in one traversal. DeepSeek V4 Pro takes a fundamentally different approach: a pipeline of six \texttt{String.replaceAll} calls, each handling one concern: camelCase splitting via \texttt{(?<=[a-z])(?=[A-Z])} lookbehind/lookahead, lowercase conversion, digit-boundary splitting, non-alphanumeric replacement, hyphen collapsing, and leading/trailing stripping. The two implementations share almost no tokens (similarity 0.124) yet produce identical output on all test cases. This illustrates why \textbf{DevBench} reports both Pass@1 and similarity metrics: functional correctness alone would treat these as equivalent, while similarity reveals they represent genuinely different problem-solving strategies: imperative single-pass parsing versus declarative multi-pass regex transformation.

\begin{center}
\textbf{Example 11: Java Pattern Matching \#20}
\end{center}

\begin{lstlisting}[style=javastyle]
public class PMTask20 {
    // Custom slugify: lowercase, non-alnum -> hyphen, collapse hyphens,
    // strip leading/trailing hyphens.
    // ADDITIONALLY: insert hyphen at camelCase boundaries (lower->Upper)
    // AND at letter<->digit boundaries.
    public static String slugify(String input) {
\end{lstlisting}

\begin{lstlisting}[style=golden, language=Java]
        StringBuilder sb = new StringBuilder();
        for (int i = 0; i < input.length(); i++) {
            char ch = input.charAt(i);
            if (!Character.isLetterOrDigit(ch)) { if (sb.length() > 0 && sb.charAt(sb.length() - 1) != '-') sb.append('-'); continue; }
            if (sb.length() > 0 && sb.charAt(sb.length() - 1) != '-' && i > 0) {
                char p = input.charAt(i - 1);
                if ((Character.isLowerCase(p) && Character.isUpperCase(ch)) || (Character.isLetter(p) && Character.isDigit(ch)) || (Character.isDigit(p) && Character.isLetter(ch))) sb.append('-');
            }
            sb.append(Character.toLowerCase(ch));
        }
        String s = sb.toString(); while (s.startsWith("-")) s = s.substring(1); while (s.endsWith("-")) s = s.substring(0, s.length()-1);
        return s;
\end{lstlisting}

\renewcommand{\modelname}{DeepSeek V4 Pro}
\begin{lstlisting}[style=model, language=Java]
        String s = input;
        // Insert hyphen at camelCase boundaries (lowercase followed by uppercase)
        s = s.replaceAll("(?<=[a-z])(?=[A-Z])", "-");
        // Convert to lowercase
        s = s.toLowerCase();
        // Insert hyphen at letter-digit boundaries
        s = s.replaceAll("(?<=[a-z])(?=\\d)|(?<=\\d)(?=[a-z])", "-");
        // Replace non-alphanumeric characters with hyphens
        s = s.replaceAll("[^a-z0-9]+", "-");
        // Collapse multiple hyphens
        s = s.replaceAll("-+", "-");
        // Strip leading/trailing hyphens
        s = s.replaceAll("^-|-$", "");
        return s;
\end{lstlisting}

\begin{lstlisting}[style=suffix, language=Java]
    }

    public static void main(String[] args) {
        // Custom slugify with camelCase and digit boundary splitting
        assert slugify("helloWorld").equals("hello-world") : "camelCase";
        assert slugify("item3pack").equals("item-3-pack") : "digit boundaries";
        assert slugify("hello world").equals("hello-world") : "spaces to hyphens";
        assert !slugify("item3pack").equals("item3pack") : "digits must be split";
        assert !slugify("helloWorld").equals("helloworld") : "camelCase must split";
        assert slugify("  leading  ").equals("leading") : "strip leading/trailing";
        System.out.println("OK: " + slugify("myVar2Go"));
    }
}
\end{lstlisting}

\section{Limitations, assumptions, and broader impacts}
\label{sec:limitations_assumptions_impacts}

\subsection{Limitations and future directions}
\label{sec:limitations}

While \textbf{DevBench} represents a significant advancement in code generation evaluation, we identify two principal directions for future work.

\subsubsection{Enhancing evaluation frameworks}

The complementary evaluation metrics we employ (Pass@1, similarity-based metrics, and LLM-judge assessments) provide multidimensional insights into model performance. The occasional divergence between these metrics, such as cases where higher syntactic similarity does not correlate with functional correctness, highlights an opportunity to develop composite metrics that better capture the full spectrum of code quality dimensions relevant to developers. Future work could also explore ensemble judging approaches combining multiple judge models, or contrastive evaluation techniques that specifically control for stylistic biases.

\subsubsection{Broadening coverage scope}

\textbf{DevBench} currently covers six major programming languages and six task categories focused on code completion. Future extensions could apply our telemetry-driven methodology to additional development activities such as code refactoring, debugging, and multi-file architecture design, as well as incorporate emerging languages such as Rust, Go, and Swift.

\subsection{Assumptions}
\label{sec:assumptions}

The evaluative claims supported by \textbf{DevBench} rest on the following assumptions:

\begin{enumerate}
\item \textbf{Fill-in-the-middle evaluation mode.} Models are evaluated using a structured FIM prompt (Appendix~\ref{subsec:evaluation_prompt}) that presents prefix, a \texttt{\#TODO} marker, and suffix. Results may differ under chat-style or agentic prompting, and scores should not be interpreted as measuring general coding ability outside this completion paradigm.

\item \textbf{Sampling representativeness.} We report Pass@1 from $n{=}5$ samples at temperature~0.2 (or default reasoning settings for models that do not expose temperature). Different sampling parameters could shift absolute scores, though we expect relative model rankings to be robust based on prior work showing rank stability across temperature settings~\citep{chenEvaluatingLargeLanguage2021}.

\item \textbf{Assertion coverage.} Functional correctness is determined by a finite set of hidden assertions per task. These assertions test key behaviors but do not guarantee full specification coverage; a completion that passes all assertions may still contain latent bugs, and one that fails may be correct on dimensions the assertions do not cover. This is precisely why \textbf{DevBench} employs a multi-metric evaluation framework: similarity-based metrics capture syntactic closeness to the golden completion regardless of assertion outcomes, and LLM-judge assessments evaluate contextual relevance and helpfulness without relying on execution. Together, the three complementary metrics mitigate the limitations of any single measure.

\item \textbf{LLM-judge scope.} The Gemini~2.5 Flash judge evaluates relevance and helpfulness without executing code. Its scores complement but do not substitute for functional correctness; systematic failure modes are documented in Appendix~\ref{sec:judge_failure_analysis}.
\end{enumerate}

\subsection{Broader impacts}
\label{sec:broader_impacts}

Our \textbf{DevBench} benchmark has several potential positive societal impacts. By enabling more accurate evaluation of code completion models, our work can lead to improved developer productivity tools that reduce repetitive coding tasks, decrease the time required to implement software solutions, and potentially lower barriers to entry in programming by assisting novice developers. More accurate code completion could also improve software quality by suggesting well-tested patterns and reducing common programming errors, potentially leading to more reliable and secure software systems.

However, we also acknowledge several potential negative impacts. First, there are fairness considerations related to programming language representation; our benchmark's coverage of six languages, while broader than many existing benchmarks, still represents a limited subset of the programming ecosystem. This may lead to uneven improvements across programming languages, potentially disadvantaging developers who work primarily with languages not included in our benchmark. Second, there are potential job market implications if increasingly capable code completion systems begin to automate significant portions of software development tasks, potentially affecting employment opportunities for certain types of programming roles.

Additionally, we recognize that improvements in code generation capabilities could have security implications. While our benchmark focuses on code completion rather than full program generation, advances in code synthesis could potentially be misused to generate malicious code more efficiently or to exploit vulnerabilities in existing systems. To mitigate these concerns, we have designed our benchmark to emphasize proper API usage, security patterns, and code quality metrics rather than merely measuring functional correctness.

To address these concerns, we have made our benchmark and methodology publicly available to enable community scrutiny, external validation, and continuous improvement. We encourage future research to extend language coverage, develop more diverse evaluation metrics, and carefully monitor potential misuses of increasingly capable code generation systems.

\end{document}